\documentclass[12pt]{fairmeta}

\usepackage{amsmath}
\usepackage{amssymb}
\usepackage{mathtools}
\usepackage{amsthm}
\newtheorem{definition}{Definition}[section]
\usepackage{bm}
\usepackage{algorithm}
\usepackage{algpseudocode}
\usepackage{paralist}
\usepackage[most]{tcolorbox}
\usepackage{booktabs}
\usepackage{arydshln} 
\usepackage{array}
\usepackage{tabularx}
\usepackage{ragged2e}
\usepackage{makecell}
\usepackage{multirow}
\usepackage{tikz}
\usepackage{pgfplots}
\usepackage{wrapfig}
\usepackage{booktabs}
\usepackage{float} 
\usepackage{longtable} 
\usepackage{xurl} 
\usepackage{fontawesome5}

\usetikzlibrary{pgfplots.groupplots, matrix}
\pgfplotsset{compat=1.18}
\usepackage{enumitem}
\usepackage{comment}
\usepackage{graphicx}
\usepackage{multirow}
\usepackage{stackengine}
\usepackage{colortbl} 
\usepackage{anyfontsize}
\newcolumntype{C}[1]{>{\centering\arraybackslash}p{#1}}

\definecolor{purple}{HTML}{c994c7}
\colorlet{navyblue}{metablue}
\colorlet{citecolor}{metablue}
\definecolor{lightgray}{gray}{0.9}
\definecolor{blanchedalmond}{rgb}{1.0, 0.92, 0.8}
\definecolor{cerise}{rgb}{0.871, 0.192, 0.388}
\definecolor{morandiblue}{HTML}{6F8191}
\definecolor{morandired}{HTML}{8F6D72}
\definecolor{metalinkpink}{HTML}{D94888}

\definecolor{TaskBG}{HTML}{EFE6FF}        
\definecolor{StateBG}{HTML}{F5F5F7}       
\definecolor{ExpertBG}{HTML}{EAF7EA}      
\definecolor{IWMBG}{HTML}{FDECF3}         
\definecolor{SRBG}{HTML}{E6F2FF}          

\newtcolorbox{trainingexample}[2][]{%
  enhanced, breakable, colframe=black!12, colback=white, boxrule=2.5pt,
  arc=2pt, left=0pt, right=0pt, top=0pt, bottom=0pt,
  title={#2}, fonttitle=\bfseries, coltitle=black, #1}

\newcolumntype{L}[1]{>{\RaggedRight\arraybackslash}p{#1}} 
\newcolumntype{Y}{>{\RaggedRight\arraybackslash}X}        

\usepackage{booktabs}
\usepackage{bbding}   
\usepackage{pifont}   
\usepackage{natbib}   

\newcommand{\cmark}{\textcolor{green!60!black}{\ding{51}}} 
\newcommand{\xmark}{\textcolor{red!70!black}{\ding{55}}}   

\usepackage[utf8]{inputenc}
\usepackage{geometry}
\usepackage{tcolorbox}
\colorlet{lightyellow}{metabg!70!white}
\colorlet{takehomerule}{metablue!70!black}
\newcounter{takenum}

\newcommand{\takeawaybullet}{\textcolor{metablue!82!black}{\raisebox{0.14ex}{\tiny$\blacktriangleright$}}}
\newtcolorbox{surveytakeawaybox}[1][]{%
  enhanced,
  breakable,
  colback=white,
  colframe=metablue!42!black,
  boxrule=0.45pt,
  borderline east={2pt}{0pt}{metablue!78!black},
  arc=2pt,
  left=7pt,
  right=7pt,
  top=5pt,
  bottom=4pt,
  boxsep=0pt,
  before skip=7pt plus 2pt minus 1pt,
  after skip=8pt plus 2pt minus 1pt,
  coltitle=metablue!85!black,
  fonttitle=\bfseries\sffamily\footnotesize,
  fontupper=\small,
  attach boxed title to top left={xshift=4pt,yshift*=-\tcboxedtitleheight/2},
  boxed title style={
    colframe=metablue!40!black,
    colback=metabg!78!white,
    boxrule=0.45pt,
    arc=2pt,
    left=5pt,
    right=5pt,
    top=1.8pt,
    bottom=1.8pt
  },
  #1
}
\newenvironment{surveytakeaways}[1][Key Takeaways]
{%
  \begin{surveytakeawaybox}[title={#1}]%
  \begin{itemize}[
    leftmargin=1.2em,
    label=\takeawaybullet,
    itemsep=1.8pt,
    topsep=2pt,
    parsep=0pt,
    partopsep=0pt
  ]
}
{%
  \end{itemize}
  \end{surveytakeawaybox}
}

\AtBeginEnvironment{thebibliography}{\sloppy\setlength{\emergencystretch}{3em}}

\makeatletter
\newcommand\fair@authsep{, }
\renewcommand\author[2][]{%
  \addtolist[#1]{#2}{\authorlist}{\authorformat}{\fair@authsep}%
  \gdef\fair@authsep{, }
}
\newcommand\authorbreak{%
  \g@addto@macro\authorlist{,\\} 
  \gdef\fair@authsep{ }          
}
\makeatother

\makeatletter
\def\fair@emails{} 
\newcommand{\emails}[1]{\gdef\fair@emails{#1}}

\pretocmd{\mymaketitle}{%
  \apptocmd{\affiliationlist}{%
    \ifdefempty{\fair@emails}{}{\\[0.35em]{\small\color{metablue}\texttt{\fair@emails}}}%
  }{}{}%
}{}{}
\makeatother

\makeatletter
\renewcommand\authorformat[2][]{%
  {\sffamily\bfseries #2$^{#1}$\@thanks}%
}
\makeatother

\renewcommand\metadataformat[2][]{ {\footnotesize {\sffamily \bfseries #1:} #2} }
\newcommand{\metalink}[2]{\href{#1}{\textcolor{metalinkpink}{\texttt{#2}}}}

\definecolor{lastauthor}{RGB}{143, 68, 115}

\title{Detecting AI-Generated Video: A Vision-Language Dual-View Survey}

\author[1]{Dylan Xinming Hou}
\author[2]{Juntian Zhang}
\author[2]{Xu Gu}
\authorbreak
\author[3]{Yichen Wu}
\author[1]{Nils Lukas}
\author[1]{Gus Xia}
\author[1]{Xiuying Chen}
\author[1\dagger]{Yuhan Liu}
\affiliation[1]{MBZUAI}
\affiliation[2]{Gaoling School of Artificial Intelligence, Renmin University of China}
\affiliation[3]{Harvard University}

\emails{dyxhou@gmail.com,
\{nils.lukas,  gus.xia, xiuying.chen, yuhan.liu\}@mbzuai.ac.ae, \\
\{zhangjuntian, guxu\}@ruc.edu.cn, %
yiwu6@mgh.harvard.edu}

\abstract{
The evolving realism of AI-generated Videos (AIGC-V) is rapidly rendering traditional artifact-centric detection insufficient, necessitating a paradigm shift from low-level inspection to high-level semantic verification.
This paper presents a comprehensive survey of AIGC-V detection, reframing the task as Factual Fidelity Verification, which asks whether the events, entities, and physical processes depicted in a video are consistent with real-world facts.
To systematize this rapidly evolving field, we propose a \textit{\textbf{Vision-Language Dual-View}} taxonomy that organizes existing methods into a hierarchical, four-layer landscape, spanning intrinsic cue analysis, spatiotemporal consistency modeling, cross-modal consistency reasoning, and language-guided world-level reasoning.
This dual-view framing highlights a fundamental transition from artifact matching in traditional deepfake detection to evidence-based semantic verification enabled by vision-language models and agentic reasoning pipelines.
Based on a systematic review of 221 works, we synthesize AIGC-V generation paradigms, survey the landscape of detection methods, and review evaluation metrics and benchmarks in line with proposed views. Finally, we discuss current challenges and identify promising directions toward robust, explainable, and trustworthy detection.
}

\date{\today}

\metadata[\faGithub\ Github]{\metalink{https://github.com/dxhou/AI-Generated-Video-Detection}{\textbf{https://github.com/dxhou/AI-Generated-Video-Detection}}}
\metadata[\faHome\ Homepage]{\metalink{https://aigcvdetection.github.io/}{\textbf{https://aigcvdetection.github.io/}}}

\begin{document}

\maketitle
\footnotetext{$\dagger$Corresponding author: Yuhan Liu\texttt{(yuhan.liu@mbzuai.ac.ae)}}
\vspace{-0.2em}
\section{Introduction}

The rapid evolution of video generation models, exemplified by Sora 2~\citep{OpenAI2024Sora2},  Veo 3~\citep{Google2025Veo3}, and Seedance 2.0~\citep{ByteDance2026Seedance2}, is fundamentally reshaping the information landscape. Unlike early deepfakes dominated by localized face swapping~\citep{wang2024fuseanypart}, modern \textbf{AI-Generated Content-Videos (AIGC-V)} have achieved cinematic fidelity with coherent narratives, increasingly blurring the boundary between synthesized fiction and captured reality. This technological leap destabilizes the foundational trust in video evidence, traditionally relied upon to verify "who, when, where, and what"~\citep{ho2022imagenvideo}.

\begin{figure}[t]
    \centering
    \includegraphics[width=1\linewidth]{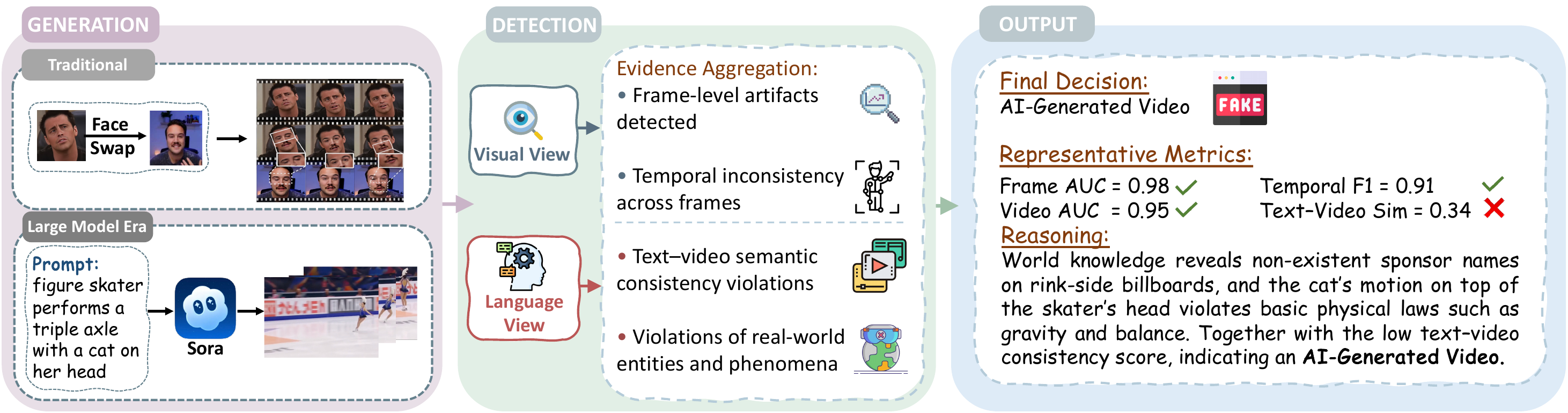}
    \caption{An example of the AIGC-V detection pipeline under our approach, illustrating AI-generated videos from traditional methods or text-to-video prompts, detection from the visual and language views, and outputs at different levels.}
    \label{fig:intro}
\end{figure}

As generation paradigms shift from local manipulation to end-to-end synthesis, such as text-to-video, traditional detection methods~\citep{wang2025dynamicface,ma2025sayanything} face a critical bottleneck. Early detection systems primarily relied on low-level visual artifacts, such as blending boundaries. However, advanced diffusion models and transformers~\citep{ho2022videodiffusion, OpenAI2024Sora} can now produce visually high-fidelity videos. This paradigm shift necessitates a new detection landscape, shifting from perceptual inspection (checking for visual artifacts) to cognitive reasoning (checking for semantic and factual violations). The surge of Vision-Language Models (VLMs)\citep{zhang2025viper,zhang2025weaving,wang2026forest} and agentic frameworks\citep{liu2024skepticism,liu2025stepwise,liu2025truth,hou2024coactgloballocalhierarchyautonomous} offers a promising pathway, enabling detectors to perform world-level reasoning and verify cross-modal consistency. Although \citet{Fu2025DeeptraceReward} and \citet{Park2025VidGuardR1} have begun to explore these directions, the existing literature remains fragmented. Meanwhile, surveys like \citet{Pei2024BenchmarkSurvey} and \citet{Liu2024EvolvingMultimodal} are constrained to the early era of deepfakes or treat video merely as a sequence of images, failing to systematize the emerging class of methods that leverage language for semantic verification.

To address this gap, we propose a \textbf{\textit{Vision-Language Dual-View}} taxonomy to organize the AIGC-V detection landscape, as illustrated in Figure~\ref{fig:intro}. We reframe the problem as \textit{Factual Fidelity Verification}, determining whether the video content aligns with real-world facts and physical laws, rather than simple binary classification. Our framework categorizes detection methods into a four-layer hierarchy, progressing from low-level perception to high-level cognition: (1) Intrinsic Cue Analysis, (2) Spatiotemporal Consistency, (3) Cross-Modal Consistency, and (4) Language-Guided World-Level Reasoning.

In this survey, we comprehensively review 221 works as of March 2026. We begin with AIGC-V paradigms (\S\ref{sec:paradigms}), then formulate AIGC-V detection from the perspective of \textit{factual fidelity} and introduce our Vision-Language dual-view framing (\S\ref{sec:detection}). We systematically organize methods under this landscape (\S\ref{sec:methods}), revisit evaluation metrics (\S\ref{sec:eval-metrics}) from a dual-view, and review benchmarks in line with AIGC-V paradigms (\S\ref{sec:benchmarks}). Finally, we identify the critical challenges (\S\ref{sec:challenges}) in detecting increasingly sophisticated AIGC-V. By bridging the gap between traditional visual forensics and emerging multimodal reasoning, this paper aims to provide a structured roadmap for the next generation of explainable and trustworthy AIGC-V detection.

\FloatBarrier

\section{Paradigms of AI-Generated Video}
\label{sec:paradigms}
\begin{figure}[t]
    \centering
    \includegraphics[width=1\linewidth]{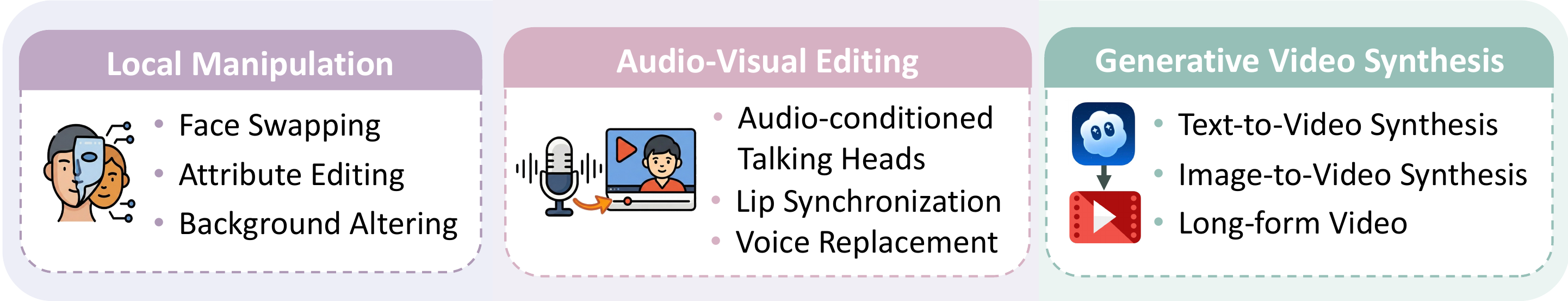}
    \caption{Overview of three AIGC-V paradigms defined in this survey, highlighting typical forms.}
    \label{fig:paradigms}
\end{figure}

Currently, diffusion-based and Transformer-based architectures are advancing rapidly in video generation, driving major progress in text-to-video and other AIGC-V pipelines~\citep{ho2022videodiffusion,ho2022imagenvideo,singer2022makeavideo,OpenAI2024Sora2,Google2025Veo3,ByteDance2026Seedance2}. In this survey, we group mainstream methodologies into three categories according to the underlying generation paradigm: local manipulation, audio-visual editing, and generative video synthesis, following representative systems in each regime~\citep{brison2025fakeparts,cheng2022videoretalking,ho2022videodiffusion}. Figure~\ref{fig:paradigms} provides an overview.

\subsection{Local Manipulation}

Methodologies categorized as \textit{Local Manipulation} typically operate on authentic video sequences by modifying specific spatial regions or distinct semantic attributes~\citep{brison2025fakeparts,heo2025fakechain}. The resulting generation maintains a high degree of structural fidelity to the original footage. 
Another technical paradigm centers on video face swapping and facial element manipulation. Representative approaches typically utilize diffusion models~\citep{wang2024fuseanypart}, 3D facial priors~\citep{wang2025dynamicface}, and controllable conditional encoding, integrating identity feature transfer with expression, pose, and illumination reconstruction from source videos into a unified generative framework. 
Given their ease of implementation and covert nature, these techniques represent a critical threat in real-world adversarial scenarios.



\subsection{Audio-Visual Editing}
\label{sec:AIGC-V_av_editing}

\textit{Audio-Visual Editing} uses speech as an explicit control signal to drive talking-head generation or to re-dub existing videos, requiring tight cross-modal alignment between audio and face dynamics while preserving identity and background ~\citep{cheng2022videoretalking}. 
And recent diffusion-based or 3D-aware approaches~\citep{ma2025sayanything,hong2025actalker} move toward unified conditional generation, where audio guidance and identity constraints are jointly modeled to produce lip-synced~\citep{li2024ditto}, visually consistent outputs. 
Due to their low operational barrier and strong perceptual plausibility, audio-visual edits can convincingly ``bind'' forged visual performances to plausible voice tracks, amplifying risks in impersonation and misinformation.


\subsection{Generative Video Synthesis}
\label{sec:AIGC-V_generation}

\textit{Generative Video Synthesis} targets end-to-end creation of complete video sequences from text or noise, fabricating both appearance and dynamics without authentic carriers. Diffusion-based generators extend image diffusion with temporal modeling for coherence, exemplified by Video Diffusion Models~\citep{ho2022videodiffusion} and cascaded high-fidelity pipelines such as Imagen Video~\citep{ho2022imagenvideo}; Make-A-Video~\citep{singer2022makeavideo} further reduces reliance on paired text-video data by leveraging text-image priors and unpaired videos.
Recent architectures further optimize representations: Show-1~\citep{zhang2025show} hybridizes pixel-based generation with latent refinement for efficiency, while Grid Diffusion~\citep{lee2024grid} unifies spatiotemporal dimensions into a single 2D grid to simplify modeling.
Transformer-style synthesis tokenizes videos and learns long-horizon structure via autoregressive or masked modeling, while DiT further highlights the scalability of Transformer backbones for diffusion-based generation~\citep{Peebles2023DiT}. At the industrial frontier, large-scale text-to-video systems such as Sora 2~\citep{OpenAI2024Sora2}, Seedance 2.0~\citep{ByteDance2026Seedance2}, and Veo 3~\citep{Google2025Veo3} demonstrate strong open-domain capability and raise urgent needs for provenance and verification.

\begin{surveytakeaways}[Where the Evidence Lives]
  \item \textbf{LMV keeps a real carrier,} so the strongest clues are usually residual and spatially localized rather than scene-wide.
  \item \textbf{AVE is constrained by coupling,} making synchrony, speaker consistency, and speech-conditioned facial motion more decisive than generic image artifacts.
  \item \textbf{GVS removes the carrier altogether,} shifting detection away from local edit traces toward long-range coherence, factual verification, and provenance.
\end{surveytakeaways}


\FloatBarrier

\begin{figure}[t]
	    \centering
	    \includegraphics[width=\linewidth]{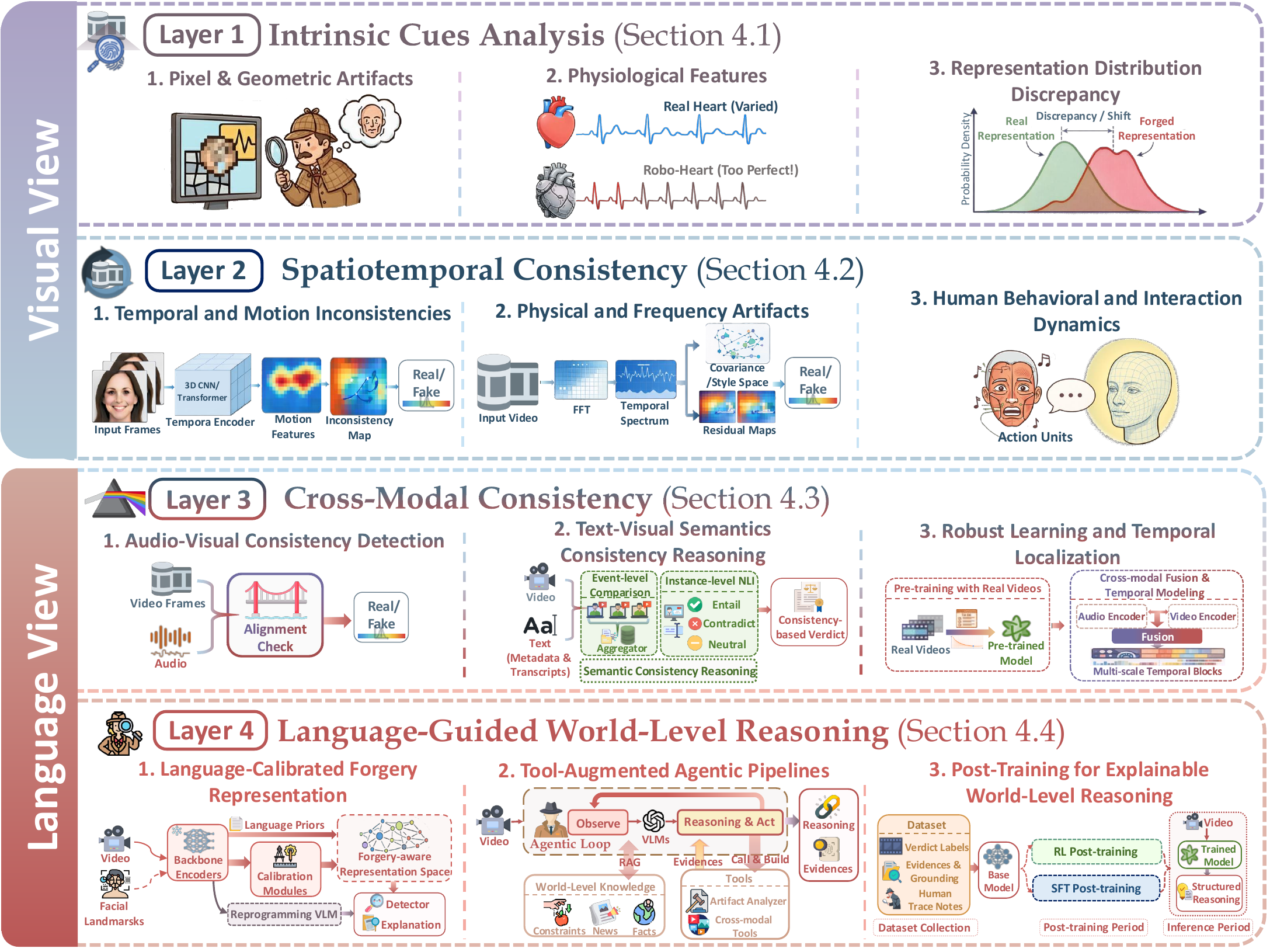}
	    \caption{Dual-view, four-layer framework for AIGC-V detection. The visual view, spanning Layers~1-2, models intrinsic cues and spatiotemporal consistency, whereas the language view, spanning Layers~3-4, addresses cross-modal consistency and world-level reasoning, with each layer further decomposed into fine-grained subcategories. Boundary note: Layer~1 uses frame-level distributional evidence; Layer~2 models frame-sequence relations; Layer~3 performs within-video multimodal verification; Layer~4 requires external knowledge for verification.}
	    \label{fig:dual_view}
\end{figure}

\section{AI-Generated Video Detection}
\label{sec:detection}

\subsection{Task Scope}
\label{sec:scope}
Visually high fidelity AIGC-V with richer narratives make the reliability of methods that output only a “real/fake” probability increasingly low in security and privacy scenarios~\citep{Wang2024DeepfakeReliability,Kaur2024DeepfakeVideoAIRE}, and the detection task should shift from merely answering the traditional question of “whether the video is AI generated” to \textbf{fact-level} judgment of whether the video content “objectively happened in the real world.”

From a fact-level perspective, video content can be abstracted as a series of propositions of the form “at a certain time and place, which entities exist, and what happens to these entities.” We define \textbf{Factual Fidelity} as “the fact-level propositions implied by the video and its metadata are consistent with the real world,” and treat \textit{Factual Fidelity Verification} as the \textbf{fundamental objective of AIGC-V detection}. 

Furthermore, different evidence pathways provided by detection methods are integrated to indicate at which levels the video violates factual fidelity~\citep{zhang2024common,Yu2025LVLMDFD,Shen2025AuthGuard} and form an interpretable and trustworthy evidence space.

\begin{definition}[Factual fidelity verification]
Let $x\in\mathcal{X}$ be a video sample together with its metadata, and let
$\Pi(x)\subseteq\mathcal{P}$ denote the set of fact-level propositions implied by $x$.
Define the ideal factual-violation label as
$
y(x)=\mathbf{1}\!\left[\Pi(x)\nsubseteq \mathcal{P}_{\mathrm{world}}\right],
$
i.e., $y(x)=1$ if and only if $x$ violates factual fidelity.
\end{definition}

\begin{definition}[AIGC-V detection]

An AIGC-V detector is a mapping
$
D:\mathcal{X}\to [0,1]\times\mathcal{E}
$. For a video sample $x\in\mathcal{X}$, 
$D(x)$ contains a \emph{suspicion score} $s(x)\in[0,1]$ 
and \emph{evidence} $e(x)\in\mathcal{E}$ supporting its judgment.

\end{definition}

We require $e(x)$ to be \emph{trustworthy} and \emph{grounded}. Concretely, we recommend representing
$e(x)$ as a finite set of evidence items $e(x)=\{\epsilon_k\}_{k=1}^K$, where each
$
\epsilon_k=(g_k, c_k, \pi_k, r_k)
$
contains: (a) a grounding $g_k$, such as temporal segment and optional spatial track, (b) a confidence $\pi_k\in[0,1]$, 
(c) a possible human-readable claim-check statement $c_k$ describing what sub-claim is being verified and what inconsistency is found
, and (d) a provenance record $r_k$ such as tool outputs or retrieved facts enabling traceability.

\subsection{Dual-view Four-Layer Taxonomy}

The methodological perspectives of AIGC-V detection can be summarized as two complementary scientific views: visual and language.
Out of this observation, as illustrated in Figure~\ref{fig:dual_view}, we propose a \textbf{Vision-Language Dual-View} approach and organize existing methods into a four-layer methodological landscape from low-level perception to high-level cognition in \S\ref{sec:methods}.

\begin{definition}[Multimodal video sample]
Let $\mathcal{I}\subseteq\mathbb{R}^{H\times W\times C}$ denote the space of frames, and let $\mathcal{A}$, $\mathcal{S}$, $\mathcal{M}$ denote the spaces of audio, text, and metadata.
A multimodal video sample is
\[
  x=(V,A,S,M)\in\mathcal{X}:=\mathcal{I}^{T}\times\mathcal{A}\times\mathcal{S}\times\mathcal{M},
\]
where $V=(I_t)_{t=1}^{T}$ is the visual stream, and $A,S,M$ may be empty when the corresponding modality is unavailable.
\end{definition}

\paragraph{\textbf{Visual View.}}
The visual view focuses on the visual modality and emphasizes statistical differences between AIGC-V and real videos~\citep{zhou2024freqblender,gu2021spatiotemporal}.
The detection pathway extends from low-level intrinsic cue analysis at the frame level in \S\ref{sec:layer1} to spatiotemporal consistency across frames in \S\ref{sec:layer2}, forming the perceptual perspective for factual-fidelity verification.

\paragraph{\textbf{Language View.}}
In this survey, the language view denotes a \emph{grounded verification pathway}, not an assumption that all audio cues are linguistic by default.
Concretely, audio evidence enters this pathway in two forms.
The first form is non-linguistic cross-modal perceptual evidence, which mainly checks within-video alignment without semantic parsing, such as lip--speech synchrony and onset and offset consistency~\citep{1a23bed3dbff4e9cb0984b74aff3376a,Yang2021PreventingDA}.
The second form is language-grounded speech evidence: speech can be mapped via ASR or speech representations to phoneme, word, or semantic units, enabling consistency checks against identity and text content, such as voiceprint--identity coherence and speech--subtitle agreement~\citep{9980296,Bohacek_2024_CVPR,cheng2022voicefacehomogeneitytellsdeepfake,cafvd2025}.
This design choice clarifies why audio can legitimately participate in the language view: it provides an interface from acoustic signals to language-level evidence and textual reasoning when needed.
Building on these two cue types, methods progress from cross-modal consistency analysis (\S\ref{sec:layer3}) to world-level reasoning at factual and knowledge levels when external evidence is required (\S\ref{sec:layer4})~\citep{zhang2024common,Yu2025LVLMDFD,Shen2025AuthGuard}, thereby forming the cognitive perspective of factual-fidelity verification.

\begin{surveytakeaways}[What AIGC-V Detection Must Establish]
  \setlength{\itemsep}{1pt}
  \setlength{\topsep}{1pt}
  \footnotesize
  \item \textbf{The target is factual fidelity, not only a real/fake score:} detection should judge whether the fact-level claims implied by a video and its metadata remain consistent with the real world.
  \item \textbf{A detector should return evidence, not only suspicion:} the output should couple a score with grounded, traceable evidence showing which claim is being checked, where the inconsistency lies, and why the decision is reliable.
  \item \textbf{Evidence can arrive through two complementary pathways:} the visual view tests intrinsic and temporal plausibility, while the language view moves from within-video cross-modal verification to explicit claim checking and world-level reasoning.
\end{surveytakeaways}

\FloatBarrier

\begin{figure}[t]
    \centering
    \includegraphics[width=\linewidth,height=0.45\textheight,keepaspectratio]{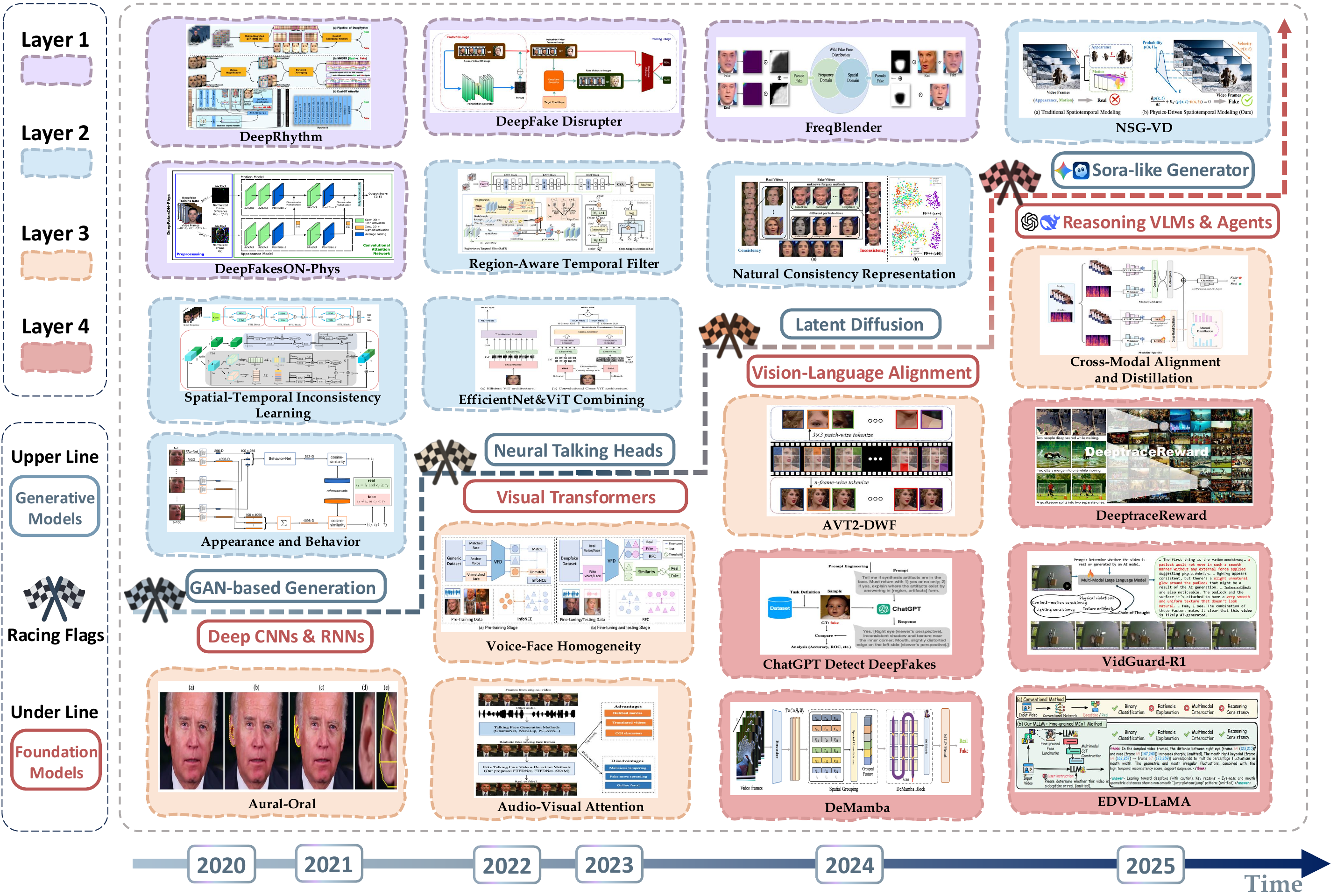}
    \caption{Landscape of representative AIGC-V detection methods aligned with the four-layer taxonomy in \S\ref{sec:methods}. Methods are grouped by the evolving capabilities of generative models that pose emerging threats and the corresponding advances in foundation models that enable detection. The year-wise shift toward language-view methods across the surveyed detection papers is quantified in Table~\ref{tab:yearly-layer-share}.}
    \label{fig:landscape}
\end{figure}

\section{Landscape of Detection Methods}
\label{sec:methods}

As illustrated in Figure~\ref{fig:landscape}, we organize AIGC-V detection methods into four layers, and provide a structured overview in this section.
Our taxonomy is \textbf{evidence-oriented rather than task-exclusive}: layer assignment follows the dominant evidence pathway used for decision-making, not a hard partition of tasks.
To make layer assignment operational, we use boundary criteria tied to evidence dependency: Layer~1 captures frame-level distributional cues that can be scored per frame and pooled, whereas Layer~2 requires explicit frame-sequence modeling of spatiotemporal relations.
Layer~3 then moves from visual-only coherence to within-video multimodal verification, whereas Layer~4 further verifies claims against external facts, commonsense, or physics, even when within-video modalities appear mutually consistent.
When a method combines cues from multiple layers, we map each module to its primary evidence type rather than forcing a single-task label.
For each layer below, we further summarize the assumption-signal-failure-mode linkage to clarify why each pathway works, when it breaks, and how recent methods improve robustness under the factual-fidelity framing in Section~\ref{sec:detection}.
To make this trend concrete, Table~\ref{tab:yearly-layer-share} summarizes the surveyed detection papers in Section~\ref{sec:methods} using complete publication years through 2025.

\subsection{Layer 1: Intrinsic Cues Analysis}
\label{sec:layer1}
Layer~1 asks whether low-level visual signals obey the statistical regularities of real videos, whether they match acquisition and post-processing effects~\citep{le2023quality,wang2022deepfake}, and whether AI generation or editing leaves intrinsic cues such as style patterns, model artifacts~\citep{Chen_2021_CVPR}, or deviant physiological signals~\citep{ciftci2020hearts,qi2020deeprhythm,ciftci2020fakecatcher}. Methods operate from the visual view by modeling, extracting, and amplifying these low-level signals~\citep{vahdati2024beyond}. The dominant evidence unit remains frame-level and distributional: a single frame can already yield an independent signal, while multiple frames mainly aggregate per-frame evidence. Once the decision depends on explicit inter-frame relations, the evidence moves to Layer~2.
Table~\ref{tab:layer1_intrinsic} summarizes representative methods in this layer.
\paragraph{\textbf{Pixel and Geometric Artifacts.}}
This line of work targets intrinsic traces introduced by AIGC-V synthesis and post-processing, including frequency fingerprints, texture residues, and geometric inconsistencies.
FreqBlender amplifies synthesis-related spectral cues via spectral blending and contrastive learning~\citep{zhou2024freqblender}.
Such frame-wise frequency fingerprints remain Layer~1 evidence, whereas descriptors that depend on temporal-frequency behavior belong to Layer~2.
Localized designs reduce reliance on global spectra by focusing on region-level residues, such as mask-guided localization in MagDR and texture-consistency learning in HCIL~\citep{Chen_2021_CVPR,gu2022hierarchical}.
Geometric cues complement spectral and texture cues by testing structural plausibility, as instantiated by LRNet, RAM, and head-pose inconsistency tests~\citep{sun2021improving,tian2024real,yang2019exposing}.
Following quality-aware training results~\citep{le2023quality}, these cues still degrade under compression and adaptive filtering in low-quality pipelines.

\paragraph{\textbf{Physiological Features.}}
Physiological cues treat faces as carriers of biological rhythms and micro-dynamics that are difficult to synthesize consistently.
Because many practical AIGC-V scenarios remain human-centric, especially talking heads and avatars, we keep physiological evidence as a dedicated subclass.
Recent follow-up is still limited: beyond classic rPPG pipelines, only a few works revisit local-global physiological interaction patterns~\citep{wu2024local,DAmelio2023rPPGNote}.
Heart-related signals are extracted either by mapping residual representations to heart-rate domains~\citep{ciftci2020hearts} or by learning spatiotemporal attention over rPPG-related patterns~\citep{qi2020deeprhythm}.
Higher-order heart-rate variability and cross-region coupling improve robustness under noise~\citep{hernandez2020deepfakeson,DBLP:conf/iccvw/FernandesROVSU019,lee2021study,wu2024local,stefanov2022visual,mao2021exposing,ciftci2020fakecatcher}.
Complementary evidence includes blink irregularities and identity-specific muscle dynamics~\citep{li2018ictu,DBLP:conf/iccv/CozzolinoRTNV21}.
These cues require stable facial visibility and sufficient temporal support, and remain sensitive to occlusion and aggressive re-encoding.

\paragraph{\textbf{Representation Distribution Discrepancy.}}
This line of work targets cross-generator transfer by modeling distribution gaps between authentic and synthetic videos rather than committing to a fixed artifact family.
Latent and style-space perturbations diversify synthetic styles and reduce forgery specificity~\citep{DBLP:conf/cvpr/YanLLLW24}, while latent-flow modeling captures abnormal style trajectories~\citep{choi2024exploiting}.
Curriculum learning and synthetic Blend-Fake data expose models to controlled difficulty and reduce shortcut learning~\citep{lin2024fake,cheng2024can}.
Recent follow-up stays close to low-level evidence: DFA combines foundation-model global priors with explicit local anomaly analysis to improve generalization under evolving forgery patterns~\citep{liao2026dfa}.
Discrepancy objectives separate overlapping real and fake manifolds with bounded contrastive learning and explicitly modeled appearance shifts~\citep{larue2023seeable,liu2024turns}.
At inference time, one-shot adaptation~\citep{chen2022ost}, quality-aware training~\citep{le2023quality}, and locality-aware reconstruction~\citep{du2020towards} improve robustness under shift while also surfacing region-level evidence for review.

\begin{surveytakeaways}[Takeaways for Layer 1]
  \item \textbf{Why It Works:} Low-level statistics still preserve residual capture and synthesis traces, so frame-wise forensic evidence can separate authentic from generated content.
  \item \textbf{Paradigm Fit:} Strongest in LMV; in AVE it is usually auxiliary unless the visual stream is also manipulated; in GVS, generator fingerprints help but transfer weakens across models, pipelines, and quality conditions.
  \item \textbf{Failure Modes:} Compression, transcoding, and domain shift attenuate these traces, and physiology-based cues are even more brittle because they require clear faces, stable signal quality, and favorable capture conditions.
\end{surveytakeaways}

\begin{table}[!htbp]
\centering
\fontsize{6.85}{7.7}\selectfont
{\setlength{\tabcolsep}{1.6pt}
\renewcommand{\arraystretch}{1}
\begin{tabularx}{\linewidth}{@{}>{\raggedright\arraybackslash}p{0.29\linewidth} Y c Y >{\raggedright\arraybackslash}p{0.10\linewidth} c@{}}
\toprule
\textbf{Method} & \textbf{Cue} & \textbf{Input} & \textbf{Mechanism} & \textbf{Output} & \textbf{Date} \\
\midrule
\rowcolor{metablue!12}
\multicolumn{6}{@{}l}{\textbf{\textcolor{metablue!90!black}{A. Pixel and geometric artifacts}}} \\
\addlinespace[0.04em]
\textbf{Inconsistent Head Poses}~\citep{yang2019exposing}
& 3D head-pose inconsistency
& V
& Head-pose estimation + geometry checks
& Score
& 05/2019 \\

\textbf{MagDR}~\citep{Chen_2021_CVPR}
& Localized artifacts around manipulated regions
& F+V
& Mask-guided localization + reconstruction loss
& Score+loc.
& 06/2021 \\

\textbf{HCIL}~\citep{gu2022hierarchical}
& Region-level inconsistency
& V
& Hierarchical contrastive learning
& Score
& 10/2022 \\

\textbf{NoiseDF}~\citep{wang2023noise}
& Forensic noise traces (face vs.\ bg)
& F
& Denoiser-based noise extraction + fusion
& Score
& 06/2023 \\

\textbf{FreqBlender}~\citep{zhou2024freqblender}
& Frequency-domain fingerprints
& F
& Spectral blending augmentation for frequency cues
& Score
& 12/2024 \\

\addlinespace[0.08em]
\cdashline{1-6}[2pt/2pt]
\addlinespace[0.08em]
\rowcolor{metablue!12}
\multicolumn{6}{@{}l}{\textbf{\textcolor{metablue!90!black}{B. Physiological features}}} \\
\addlinespace[0.04em]
\textbf{In Ictu Oculi}~\citep{li2018ictu}
& Eye-blink irregularities
& V
& Blink detection + temporal pattern modeling
& Score
& 12/2018 \\

\textbf{FakeCatcher}~\citep{ciftci2020fakecatcher}
& PPG-like biological signal maps
& V
& Biological signal maps + detector
& Score
& 07/2020 \\

\textbf{Hearts}~\citep{ciftci2020hearts}
& Heart-related signals in residuals
& V
& Residual maps $\rightarrow$ rPPG features
& Score
& 09/2020 \\

\textbf{DeepRhythm}~\citep{qi2020deeprhythm}
& Visual heartbeat rhythms (rPPG)
& V
& Spatiotemporal attention over rPPG patterns
& Score
& 10/2020 \\

\textbf{DeepFakesON-Phys}~\citep{hernandez2020deepfakeson}
& Heart-rate estimation inconsistency
& V
& rPPG heart-rate estimation + anomaly cues
& Score
& 10/2020 \\

\textbf{Local rPPG Interaction}~\citep{wu2024local}
& Cross-region rPPG coupling
& V
& Local attention + long-range rPPG interaction
& Score
& 02/2024 \\

\addlinespace[0.08em]
\cdashline{1-6}[2pt/2pt]
\addlinespace[0.08em]
\rowcolor{metablue!12}
\multicolumn{6}{@{}l}{\textbf{\textcolor{metablue!90!black}{C. Distribution discrepancy and robustness}}} \\
\addlinespace[0.04em]
\textbf{OST}~\citep{chen2022ost}
& Domain shift and compression sensitivity
& F
& One-shot test-time adaptation
& Score
& 12/2022 \\

\textbf{SeeABLE}~\citep{larue2023seeable}
& Soft discrepancies under shifts
& F
& Real-only bounded contrastive learning
& Score
& 10/2023 \\

\textbf{QAD}~\citep{le2023quality}
& Compression-robust representations
& F
& Quality-aware regularization
& Score
& 10/2023 \\

\textbf{LSDA}~\citep{DBLP:conf/cvpr/YanLLLW24}
& Cross-generator transfer
& V
& Latent-space augmentation for transfer
& Score
& 06/2024 \\

\textbf{Style Latent Flows}~\citep{choi2024exploiting}
& Abnormal style-latent trajectories
& V
& Style-flow modeling + contrastive loss
& Score
& 06/2024 \\

\textbf{Fake It Till You Make It}~\citep{lin2024fake}
& Curriculum-based forgery augmentation
& V
& Curriculum forgery augmentation
& Score
& 10/2024 \\

\textbf{DFA}~\citep{liao2026dfa}
& Global-local intrinsic anomalies
& F+V
& CLIP adapter + local anomaly stream
& Score
& 03/2026 \\
\bottomrule
\end{tabularx}}
\caption{\textbf{Layer 1: Intrinsic cues analysis.} Representative methods grouped by cue family. \textit{Cue} = evidence source; \textit{Mechanism} = modeling design; Input: F=frames, V=video.}
\label{tab:layer1_intrinsic}
\end{table}

\FloatBarrier

\subsection{Layer 2: Spatiotemporal Consistency}
\label{sec:layer2}
Layer~2 asks whether spatiotemporal image flow satisfies the motion constraints of real videos. Real capture is constrained by continuous camera trajectories and physical scenes~\citep{zhang2025physics,interno2025ai}, so adjacent frames exhibit continuous, predictable, and physically feasible changes, whereas AIGC-V can show long-range inconsistencies such as object or background distortion~\citep{gu2022region,zhang2024learning} and sudden local blurring~\citep{gu2022delving}. Spatiotemporal inconsistency learning~\citep{gu2021spatiotemporal} and motion-based detection~\citep{ma2024decof,yin2023dynamic,haliassos2021lips} therefore exploit frame differences and flow residuals~\citep{chen2025gc}. This makes Layer~2 distinct from Layer~1 by requiring explicit sequence modeling of inter-frame relations rather than frame-wise cue pooling. At the same time, Layer~2 remains a visual-modality layer: once verification requires semantic checks across audio, text, and visual events, the evidence pathway transitions to Layer~3.
Table~\ref{tab:layer2_spatiotemporal} summarizes representative methods in this layer.

\paragraph{\textbf{Temporal and Motion Inconsistencies.}}
Layer~2 makes explicit how temporal coherence is operationalized as a modeling prior.
Instead of classifying frames independently, spatiotemporal methods model how facial appearance and geometry evolve across time and treat incoherence as evidence for forgery~\citep{gu2021spatiotemporal,gu2022region}.
Clip encoders treat short sequences as space-time volumes and learn joint patterns with 3D CNNs or attention~\citep{gu2021spatiotemporal,gu2022region,hu2021dynamic,zhang2024learning,chen2020fsspotter,zhang2021detecting}.
By jointly observing multiple frames, they reveal abnormal changes that are weak or ambiguous at the single-frame level.
Region-aware variants highlight regions such as the eyes and mouth where features change abnormally, and auxiliary objectives can predict inconsistency maps to force localized temporal evidence~\citep{gu2022region,hu2021dynamic,zhang2024learning}.
Temporal Dropout and rhythm perturbation further reduce overfitting to specific frame rates or speaking rhythms~\citep{chen2020fsspotter,zhang2021detecting}.
To reuse 2D backbones, some methods reshape frame sequences into grid-like layouts to enable efficient 2D processing and leverage image pre-training~\citep{xu2024learning,xu2023tall}.
Hybrid pipelines aggregate per-frame features with recurrent branches to fuse static appearance and dynamic motion cues at the decision stage~\citep{masi2020two}.
Application-oriented follow-ups continue to revisit CNN or LSTM aggregation as a lightweight temporal baseline for video deepfake detection~\citep{bahadure2026deepfake}.
Motion-centric designs construct intermediate signals such as frame differences or optical flow to emphasize temporal transitions~\citep{ma2024decof,yin2023dynamic}.
Flow-residual cues test whether local facial motion is compatible with global head motion, remaining informative under blur and compression~\citep{chen2025gc,wang2023exploiting}.
Overall, spatiotemporal inconsistency methods frame deepfake detection as coherence analysis rather than static appearance classification.

\paragraph{\textbf{Human Behavioral and Interaction Dynamics.}}
Behavioral cues move from artifact matching to higher-level human realism, asking whether expressions, gaze, and identity evolve plausibly over time.
Action-unit guided and appearance-behavior joint representations model muscle activations and expression trajectories to expose abnormal combinations or unnatural dynamics~\citep{anand2025detecting,agarwal2020detecting}.
Gaze-based cues probe attention and interaction patterns that are difficult to reproduce consistently in face swapping or reenactment~\citep{kohler2025deepfake}.
Identity-driven methods test whether temporal dynamics remain compatible with the claimed person under varying pose, illumination, and compression~\citep{dong2020identity}.
Geometry-consistency checks enforce compatibility between 2D observations and 3D face and head motion, exposing pose- or shape-inconsistent sequences~\citep{sun2021improving,yang2019exposing,tursman2020towards}.
These semantics- and geometry-aware cues can be strong against visually polished fakes but are sensitive to natural behavioral diversity and recording conditions.

\paragraph{\textbf{Physical and Frequency Artifacts.}}
Transform-domain designs compute descriptors where non-physical dynamics become separable from content variation, yielding traces that can be more stable than raw RGB.
Compared with purely spatial or spatiotemporal backbones trained on RGB, these descriptors can be more robust to appearance changes and sometimes more transferable across generators~\citep{kim2025beyond}.
Temporal frequency responses expose abnormal periodic components and energy distributions along time, helping separate authentic motion spectra from synthetic artifacts~\citep{nie2024dip,kim2025beyond}.
Early-2026 work pushes this line toward higher-fidelity synthetic videos: MPF-Net combines static off-manifold deviation scoring with micro-temporal fluctuation analysis, arguing that visually polished clips can still reveal structured instability around the real-video manifold~\citep{DBLP:journals/corr/abs-2601-21408}.
Second-order statistics capture covariance mismatches and subtle distribution shifts in spatiotemporal features that survive mild post-processing~\citep{zheng2025d3,padhi2025fake}.
Residual- and difference-based features amplify deviations from real-video manifolds by emphasizing temporal changes rather than absolute appearance~\citep{xu2025vod,stamnas2025difffake}.
Physics-inspired priors provide complementary constraints on motion feasibility and perceptual plausibility, linking detection to camera trajectories and physical scenes~\citep{zhang2025physics,interno2025ai}.
Two representative detectors in this subclass are NSG-VD~\citep{zhang2025physics} and ReStraV~\citep{interno2025ai}.
NSG-VD injects explicit physics constraints into spatiotemporal modeling by matching normalized spatiotemporal gradients and distribution statistics between real and generated videos, while ReStraV scores how far latent motion trajectories deviate from straight, natural perceptual manifolds.

\paragraph{\textbf{Generalization and Robustness.}}
Beyond cue design, robustness strategies aim to maintain performance under unseen generators and post-processing by changing objectives, architectures, and training protocols~\citep{nguyen2025vulnerability,yan2025generalizing,trinh2021interpretable,coccomini2022combining,tong2025deepfake}.
Vulnerability-aware objectives and plug-and-play adaptation modules steer backbones toward transferable spatiotemporal evidence~\citep{nguyen2025vulnerability,yan2025generalizing}.
Unified or cross-model detectors reuse pretrained backbones and expand the hypothesis space to handle diverse manipulations and media types~\citep{trinh2021interpretable,veeramachaneni2025leveraging,coccomini2022combining,liu2025deepfake,tariq2021one}.
Forensic augmentation and quality-centric curricula expand coverage of perturbations and hard cases~\citep{corvi2025seeing,song2024quality}.
Self-supervised video forensics learns normal dynamics via synchronization or anomaly objectives and then treats deviations as manipulations~\citep{yang2025video,feng2023self}.
Multiple-instance formulations aggregate clip-level evidence to reduce reliance on exhaustive forged labels and handle mixed-quality videos~\citep{li2020sharp}.

\begin{surveytakeaways}[Takeaways for Layer 2]
  \item \textbf{Why It Works:} Real capture follows continuous trajectories and feasible motion, whereas generated video still accumulates subtle long-horizon temporal inconsistencies.
  \item \textbf{Paradigm Fit:} Especially useful in GVS and in LMV when edits disturb motion across frames; in AVE, it matters mainly when the visual stream itself is synthesized, such as lip reenactment.
  \item \textbf{Failure Modes:} Short clips, resampling, and compression dilute the signal, and detectors may still collapse to frame-level shortcuts instead of genuine temporal reasoning.
\end{surveytakeaways}

\begin{table}[!htbp]
\centering
\scriptsize
{\setlength{\tabcolsep}{1.6pt}
\renewcommand{\arraystretch}{1}
\begin{tabularx}{\linewidth}{@{}>{\raggedright\arraybackslash}p{0.31\linewidth} Y Y >{\raggedright\arraybackslash}p{0.10\linewidth} c@{}}
\toprule
\textbf{Method} & \textbf{Cue} & \textbf{Mechanism} & \textbf{Output} & \textbf{Date} \\
\midrule
\rowcolor{metablue!12}
\multicolumn{5}{@{}l}{\textbf{\textcolor{metablue!90!black}{A. Temporal and motion inconsistencies}}} \\
\addlinespace[0.08em]
\textbf{FS-Spotter}~\citep{chen2020fsspotter}
& Spatiotemporal swap artifacts
& Fuse spatial+temporal cues
& Score
& 07/2020 \\

\textbf{Dynamic Prototypes}~\citep{trinh2021interpretable}
& Transferable spatiotemporal evidence
& Dynamic prototypes + predictive learning
& Score
& 01/2021 \\

\textbf{Temporal Dropout 3DCNN}~\citep{zhang2021detecting}
& Frame-rate dependent temporal artifacts
& 3D CNN + temporal dropout
& Score
& 08/2021 \\

\textbf{STIL}~\citep{gu2021spatiotemporal}
& Short-clip spatiotemporal inconsistency
& Clip encoder for ST inconsistency
& Score
& 10/2021 \\

\textbf{EfficientNet-ViT Ensemble}~\citep{coccomini2022combining}
& Robust temporal features
& CNN and ViT ensemble
& Score
& 05/2022 \\

\textbf{Region-Aware Temporal Inconsistency}~\citep{gu2022region}
& Region-wise dynamics anomalies
& Region-aware learning + maps
& Score+loc.
& 07/2022 \\

\textbf{TALL}~\citep{xu2023tall}
& Cross-frame inconsistency
& Thumbnail layout + 2D backbone
& Score
& 10/2023 \\

\textbf{Natural Consistency Representation}~\citep{zhang2024learning}
& Temporal naturalness deviations
& SSL consistency + fine-tune
& Score
& 11/2024 \\

\textbf{Vulnerability-Aware Learning}~\citep{nguyen2025vulnerability}
& Cross-generator vulnerability patterns
& Vulnerability-aware transfer
& Score
& 01/2025 \\

\textbf{GC-ConsFlow}~\citep{chen2025gc}
& Optical-flow residual inconsistency
& Flow residuals + global context
& Score
& 06/2025 \\

\textbf{Plug-and-Play Adapters}~\citep{yan2025generalizing}
& Generalization under generator shift
& Video blending + ST adapters
& Score
& 06/2025 \\
\addlinespace[0.12em]
\cdashline{1-5}[2pt/2pt]
\addlinespace[0.12em]
\rowcolor{metablue!12}
\multicolumn{5}{@{}l}{\textbf{\textcolor{metablue!90!black}{B. Physical and frequency artifacts}}} \\
\addlinespace[0.08em]
\textbf{DIP}~\citep{nie2024dip}
& Temporal frequency response anomalies
& Transform-domain temporal spectra
& Score
& 12/2024 \\

\textbf{ReStraV (Perceptual Straightening)}~\citep{interno2025ai}
& Non-physical motion and trajectory irregularities
& Perceptual straightening
& Score
& 07/2025 \\

\textbf{Beyond RGB}~\citep{kim2025beyond}
& Stable descriptors beyond raw RGB
& Transformed descriptors
& Score
& 07/2025 \\

\textbf{Physics-Driven ST Modeling}~\citep{zhang2025physics}
& Physics constraints on spatiotemporal flow
& Physics-driven ST modeling
& Score
& 10/2025 \\

\textbf{MPF-Net}~\citep{DBLP:journals/corr/abs-2601-21408}
& Off-manifold residuals + micro-temporal fluctuations
& Hierarchical manifold deviation + temporal filtering
& Score
& 01/2026 \\

\addlinespace[0.12em]
\cdashline{1-5}[2pt/2pt]
\addlinespace[0.12em]
\rowcolor{metablue!12}
\multicolumn{5}{@{}l}{\textbf{\textcolor{metablue!90!black}{C. Human behavioral and interaction dynamics}}} \\
\addlinespace[0.08em]
\textbf{Emotions Don't Lie}~\citep{mittal2020emotionsdontlieaudiovisual}
& Affective audio-visual cue coherence
& Siamese audio-visual model + triplet loss
& Score
& 03/2020 \\

\textbf{Appearance and Behavior}~\citep{agarwal2020detecting}
& Appearance and behavior realism over time
& Behavior-aware temporal features
& Score
& 12/2020 \\

\textbf{Identity-Driven Detection}~\citep{dong2020identity}
& Identity dynamics consistency
& Track identity dynamics
& Score
& 12/2020 \\

\textbf{Gaze Tracking}~\citep{kohler2025deepfake}
& Gaze dynamics
& Gaze features for dyadic calls
& Score
& 09/2025 \\
\bottomrule
\end{tabularx}}
\caption{\textbf{Layer 2: Spatiotemporal consistency.} Representative methods grouped by (A) temporal and motion inconsistencies, (B) physical and frequency artifacts, and (C) human behavioral and interaction dynamics in \S\ref{sec:layer2}. \textit{Cue} = evidence source; \textit{Mechanism} = modeling design. All methods take video as input.}
\label{tab:layer2_spatiotemporal}
\end{table}

\FloatBarrier

\subsection{Layer 3: Cross-Modal Consistency}
\label{sec:layer3}

Layer~3 focuses on whether the modalities in a video are well aligned in describing the same thing. Real videos often contain audio, text, and visuals that are highly aligned across modalities, whereas AIGC-V may present mismatches between lip movements and speech or between identity and voiceprint~\citep{1a23bed3dbff4e9cb0984b74aff3376a,Yang2021PreventingDA,9428368,cheng2022voicefacehomogeneitytellsdeepfake}. Mismatches can also appear between image content and associated text, motivating text-video consistency reasoning~\citep{cafvd2025,cscl2025,t3svfnd2025}. Symbolic and semantic lip-sync analyses further enlarge this consistency space~\citep{9980296,Bohacek_2024_CVPR,li2024zeroshotfakevideodetection,koutlis2025dimodifdiscoursemodalityinformationdifferentiation}. Compared with Layer~2, the key shift is from visual-only sequence coherence to within-video multimodal verification; compared with Layer~4, evidence in Layer~3 is still primarily within-video and does not require external world knowledge as a mandatory source.

To clarify the role of audio in this layer, our taxonomy does not treat all audio cues as linguistic by default. Instead, we distinguish two cue types that naturally coexist in Layer~3. The first type is non-linguistic cross-modal perceptual cues, which test alignment without semantic parsing, such as lip-speech synchrony and onset and offset consistency. The second type is language- grounded speech cues, where speech is mapped via ASR or speech representations into phoneme, word, or semantic units, enabling explicit verification against identity and text content, such as voiceprint-identity coherence and speech-subtitle agreement, and providing a language interface for further reasoning when external knowledge is needed. 
Table~\ref{tab:layer3_crossmodal} summarizes representative methods in this layer.

\paragraph{\textbf{Audio-Visual Consistency Detection.}}
Audio-visual consistency methods decompose evidence into synchrony, symbolic alignment, and identity coherence.
Synchrony cues compare speech dynamics with mouth motion and learn cross-modal embeddings that penalize misalignment~\citep{1a23bed3dbff4e9cb0984b74aff3376a,Yang2021PreventingDA}.
Cues extend beyond the mouth to additional regions whose micro-motions correlate with speech production and head dynamics~\citep{Agarwal_2021_CVPR}.
Symbolic alignment improves interpretability by mapping audio and visual streams to token or phoneme sequences and comparing them at the sequence level~\citep{9980296,Bohacek_2024_CVPR,li2024zeroshotfakevideodetection,koutlis2025dimodifdiscoursemodalityinformationdifferentiation}.
Identity coherence shifts the question from ``what is said'' to ``who is speaking'' by testing whether voice and face embeddings belong to the same person, including person-of-interest settings trained with authentic talking-head data only~\citep{9428368,cheng2022voicefacehomogeneitytellsdeepfake,cozzolino2023audiovisualpersonofinterestdeepfakedetection,muppalla2023integratingaudiovisualfeaturesmultimodal}.
ConLLM further couples contrastive multimodal alignment with LLM-based reasoning to reduce modality fragmentation and expose fine-grained inconsistencies that are difficult to separate with shallow fusion alone~\citep{DBLP:journals/corr/abs-2601-17530}.
Generator-side probes such as X-AVDT further exploit inversion-accessed cross-attention signals to expose hidden speech-motion correspondence errors, while AV-LMMDetect shows that large multimodal models can also be directly fine-tuned as within-video audio-visual verifiers without requiring external world knowledge~\citep{kim2026xavdt,cao2026avlmm}.

\paragraph{\textbf{Text-Video Semantic Consistency Reasoning.}}
Text-video semantic consistency reasoning reframes detection as claim verification: given a video and its associated text, such as titles, captions, or OCR and ASR transcripts, the model tests whether the textual statements are supported by visual evidence, with semantic contradictions serving as a key signal.
CA-FVD supervises cross-modal consistency by prompting a VLM to obtain pairwise modality-consistency pseudo labels across visual, text, and audio streams, then training with cosine-based consistency losses; co-attention fusion and collaborative diagnosis aggregate modality-level confidence into a final score~\citep{cafvd2025}.
CSCL moves from global scoring to grounded evidence by building patch-token consistency matrices and using cascaded decoders to separately model within-modality context and cross-modality semantics; forgery-aware aggregation reduces confusing but locally consistent content and enables localization~\citep{cscl2025}.
To handle event shifts, T$^3$SVFND introduces test-time training with a reconstruction objective conditioned on multimodal context, suggesting event-robust text-video reasoning for detection settings~\citep{t3svfnd2025}.

\paragraph{\textbf{Robust Learning and Temporal Localization.}}
In realistic settings, cross-modal inconsistencies are often temporally sparse, so localization is needed not only for final scoring but also for review and diagnosis. Robust methods therefore need to remain stable under compression, language shift, and partial-stream corruption while still marking when and where the mismatch emerges. This couples robustness with interpretability: a useful detector should preserve discrimination under noise while exposing the specific temporal span that carries the cross-modal failure rather than diffusing evidence over the whole clip.
Two-stream fusion and attention-based alignment provide a common backbone for joint modeling and interpretation~\citep{9710387,wang2022audiovisualattentionbasedmultimodal}.
Transferable representations learned from large-scale authentic data improve stability under compression and language shifts~\citep{liang2025speechforensicsaudiovisualspeechrepresentation,feng2023self}.
Weak supervision with video-level labels and bias-aware objectives enables fine-grained temporal localization without dense annotations~\citep{Xu_2025,astrid2025audiovisualdeepfakedetectionlocal}.
Audio shift prediction provides an additional self-supervised proxy for sparse cross-modal disturbance and improves temporal localization in multimodal deepfake settings~\citep{anshul2026avshift}.
Challenge-driven systems such as Divide and Conquer~\citep{li2026divide} also show the value of decomposing detection into audio-side and visual-side localization before late fusion, which is particularly useful when manipulations are sparse or concentrated in only one stream.

\begin{surveytakeaways}[Takeaways for Layer 3]
  \item \textbf{Why It Works:} Authentic videos obey strong within-video coupling across modalities, so mismatches in timing, identity, or semantics are hard to hide consistently.
  \item \textbf{Paradigm Fit:} Natural fit for AVE; useful in LMV when edits change speaker identity or event semantics and usable audio or text is present; in GVS, it is strongest for captioned or prompt-conditioned clips.
  \item \textbf{Failure Modes:} Its value drops when audio or text is missing, noisy, or weakly informative, and attackers can partially restore apparent alignment through redubbing, retiming, or other resynchronization.
\end{surveytakeaways}

\begin{table}[!htbp]
\centering
\scriptsize
{\setlength{\tabcolsep}{1.6pt}
\renewcommand{\arraystretch}{1.2}
\begin{tabularx}{\linewidth}{@{}>{\raggedright\arraybackslash}p{0.29\linewidth} Y c Y >{\raggedright\arraybackslash}p{0.10\linewidth} c@{}}
\toprule
\textbf{Method} & \textbf{Cue} & \textbf{Input} & \textbf{Mechanism} & \textbf{Output} & \textbf{Date} \\
\midrule
\rowcolor{metablue!12}
\multicolumn{6}{@{}l}{\textbf{\textcolor{metablue!90!black}{A. Audio-visual consistency detection}}} \\
\addlinespace[0.08em]
\textbf{Not Made for Each Other}~\citep{1a23bed3dbff4e9cb0984b74aff3376a}
& Audio-visual dissonance
& Speech+Face
& Cross-modal mismatch + localization
& Score+loc.
& 10/2020 \\

\textbf{Dynamic Lip Movement}~\citep{Yang2021PreventingDA}
& Lip motion vs.\ speech mismatch
& Speech+Lip
& Lip-motion dynamics analysis
& Score
& 12/2020 \\

\textbf{Aural-Oral Dynamics}~\citep{Agarwal_2021_CVPR}
& Aural-oral dynamics coherence
& Speech+Lip
& Audio+mouth dynamics features
& Score
& 06/2021 \\

\textbf{Joint Audio-Visual Detection}~\citep{9710387}
& Audio-visual synchrony and semantics
& Speech+Face
& Joint synchrony+semantic modeling
& Score
& 10/2021 \\

\textbf{Audio-Visual Attention}~\citep{wang2022audiovisualattentionbasedmultimodal}
& Cross-modal alignment
& Speech+Face
& Cross-attention fusion
& Score
& 03/2022 \\

\textbf{Lip Sync Matters}~\citep{9980296}
& Symbolic lip-sync mismatch
& Speech+Lip
& Token and phoneme sequence comparison
& Score
& 11/2022 \\

\textbf{Voice-Face Homogeneity}~\citep{cheng2022voicefacehomogeneitytellsdeepfake}
& Voice-face identity coherence
& Voice+Face-ID
& Compare voice and face-ID embeddings
& Score
& 11/2023 \\

\textbf{Lost in Translation}~\citep{Bohacek_2024_CVPR}
& Language-aware audio-visual mismatch
& Speech+Lip
& Language-conditioned lip-sync
& Score
& 06/2024 \\

\textbf{AVFF}~\citep{oorloff2024avffaudiovisualfeaturefusion}
& Intrinsic audio-visual correspondences
& Speech+Face
& Real-only SSL pretrain + classifier
& Score
& 06/2024 \\

\textbf{Multi-task Audio-Visual Prompt Learning}~\citep{10.1609/aaai.v39i1.32042}
& Fine-grained audio-visual consistency
& Speech+Face
& Prompting + matching loss + fusion
& Score
& 04/2025 \\

\textbf{CAD}~\citep{du2025cadgeneralmultimodalframework}
& Semantic audio-visual misalignment + cues
& Speech+Face
& Alignment + distillation
& Score
& 05/2025 \\

\textbf{PIA}~\citep{datta2025piadeepfakedetectionusing}
& Phoneme-timing mismatch + ID dynamics
& Speech+Lip
& Phoneme timing + ID dynamics
& Score
& 10/2025 \\

\textbf{KLASSify to Verify}~\citep{kukanov2025klassifyverifyaudiovisualdeepfake}
& Robust audio-visual cues (unseen attacks)
& Speech+Face
& Audio SSL + graph attn + handcrafted
& Score+loc.
& 10/2025 \\

\textbf{ConLLM}~\citep{DBLP:journals/corr/abs-2601-17530}
& Multimodal semantic inconsistency
& Speech+Face
& Contrastive alignment + LLM reasoning
& Score
& 01/2026 \\

\textbf{X-AVDT}~\citep{kim2026xavdt}
& Generator-internal audio-visual correspondence
& Speech+Face
& DDIM inversion + cross-attention probing
& Score
& 03/2026 \\

\addlinespace[0.12em]
\cdashline{1-6}[2pt/2pt]
\addlinespace[0.12em]
\rowcolor{metablue!12}
\multicolumn{6}{@{}l}{\textbf{\textcolor{metablue!90!black}{B. Text-video semantic consistency reasoning}}} \\
\addlinespace[0.08em]
\textbf{CA-FVD}~\citep{cafvd2025}
& Video-text semantic mismatch
& Video+Text
& VLM pseudo-labels + consistency loss
& Score
& 04/2025 \\

\textbf{CSCL}~\citep{cscl2025}
& Text-image inconsistency grounding
& Frame+Text
& Cascaded consistency decoders
& Score+loc.
& 06/2025 \\

\textbf{T$^3$SVFND}~\citep{t3svfnd2025}
& Event-shifted semantics
& Video+Text
& Test-time MLM reconstruction
& Score
& 07/2025 \\

\addlinespace[0.12em]
\cdashline{1-6}[2pt/2pt]
\addlinespace[0.12em]
\rowcolor{metablue!12}
\multicolumn{6}{@{}l}{\textbf{\textcolor{metablue!90!black}{C. Robust learning and temporal localization}}} \\
\addlinespace[0.08em]
\textbf{Cross- and Within-Modality Regularization}~\citep{zou2024crossmodalitywithinmodalityregularizationaudiovisual}
& Modality separation under perturb.
& Speech+Face
& Audio-visual Transformer + cross and within regs
& Score
& 04/2024 \\

\textbf{Audio-Visual Local Inconsistencies}~\citep{astrid2025audiovisualdeepfakedetectionlocal}
& Local temporal audio-visual inconsistency
& Speech+Face
& Local inconsistency + localization
& Score+loc.
& 04/2025 \\

\textbf{Circumventing Shortcuts}~\citep{Smeu_2025_CVPR}
& Shortcut-robust audio-visual reps (silence)
& Speech+Face
& Real-only SSL audio-visual alignment
& Score
& 06/2025 \\

\textbf{SpeechForensics}~\citep{liang2025speechforensicsaudiovisualspeechrepresentation}
& Real-only audio-visual repr.\ learning
& Speech+Lip
& SSL pretrain on real + fine-tune
& Score
& 08/2025 \\

\textbf{WMMT}~\citep{xu2025weaklysupervisedmultimodaltemporal}
& Weakly-sup.\ temporal loc.
& Speech+Face
& Multitask + MoE + deviation loss
& Score+loc.
& 08/2025 \\

\textbf{HOLA}~\citep{wu2025holaenhancingaudiovisualdeepfake}
& Hierarchical audio-visual aggregation
& Speech+Face
& SSL pretrain + gated aggregation
& Score
& 10/2025 \\

\textbf{Weakly-Supervised Temporal Localization}~\citep{Xu_2025}
& Sparse multimodal deviations
& Speech+Face
& Weak supervision + deviation modeling
& Score+loc.
& 10/2025 \\

\textbf{A-V Shift Prediction}~\citep{anshul2026avshift}
& Temporal audio-visual perturbation cues
& Speech+Face
& Audio shift prediction + localization
& Score+loc.
& 01/2026 \\

\textbf{Divide and Conquer}~\citep{li2026divide}
& Cross-modal fusion + temporal localization
& Speech+Face
& Unimodal detectors + score fusion
& Score+loc.
& 02/2026 \\
\bottomrule
\end{tabularx}}
\caption{\textbf{Layer 3: Cross-modal consistency.} Representative methods grouped by (A) audio-visual alignment, (B) text-video semantics, and (C) robust learning and localization.}
\label{tab:layer3_crossmodal}
\end{table}

\FloatBarrier

\subsection{Layer 4: Language-Guided World-Level Reasoning}
\label{sec:layer4}

Layer~4 elevates detection from internal consistency within the video to consistency with world-level rules and knowledge. The research question shifts to whether the video content can truly exist in the real world and remain reasonable in semantic and factual dimensions. Real videos should be consistent with real-world facts, physical rules, domain knowledge, and basic common sense, whereas AIGC-V often fails to align fully with these constraints, creating the detection space exploited by this layer~\citep{motamed2025generative}. Language serves as a natural interface for human-readable explanations, while baseline VLM prompting in zero-shot or few-shot settings provides a minimal route for detecting inconsistencies and producing textual rationales~\citep{Jia2024CVPRW,Shahzad2025ChatGPTAV}. Compared with Layer~3, the detector is no longer asked only whether modalities agree, but whether the implied claim remains tenable once external facts, commonsense, or physics are brought into the loop.
Table~\ref{tab:l4_summary} summarizes representative methods in this layer.

\paragraph{\textbf{Language-Calibrated Forgery Representations.}}
Language-calibrated designs inject prompts or textual priors into multimodal representations to reshape forgery-sensitive features without rebuilding the backbone.
CPML aligns physiological pulse (rPPG) signals and facial landmark dynamics with prompts and enforces cross-quality and cross-modal consistency to stabilize physiological evidence under compression~\citep{Lai2024CPML,DAmelio2023rPPGNote}.
RepDFD reprograms pretrained vision-language models by freezing the backbone and learning input-side perturbations and adaptive prompts, while knowledge-guided variants build textual prototypes and uncertainty modeling for improved transfer~\citep{Lin2025RepDFD,Yu2025LVLMDFD,Shen2025AuthGuard}.
Adapter-style reprogramming provides a lightweight alternative when full fine-tuning is infeasible~\citep{shao2024deepfakeadapterdualleveladapterdeepfake}. These methods are useful when external retrieval is unavailable but language priors still need to bind visual cues to explicit semantic hypotheses and human-readable defect categories.

\paragraph{\textbf{Tool-Augmented Agentic Pipelines.}}
Tool-augmented pipelines cast AIGC-V detection as explicit evidence gathering, where language guides what to inspect and which analyzer to call next.
LAVID uses an observe-tool-integrate loop: it forms a coarse hypothesis, calls external tools, updates prompts based on intermediate outputs, and fuses multi-step evidence~\citep{Liu2025LAVID}.
FakeHunter adds memory-anchored observe-think-act routines, including retrieval of semantic anchors and reuse of intermediate results across steps~\citep{chen2025memory}.
DAVID-XR1 and related pipelines emphasize semantic anchors and modular analyzers to support interpretable evidence composition~\citep{gao2025david}.
Multi-agent coordination further decomposes modality-specific evidence before fusion~\citep{zaman2025deepagent}. The attraction of this design is that the final verdict can be traced to retrieved facts, tool outputs, and intermediate judgments rather than a single opaque forward pass.

\paragraph{\textbf{Post-Training for Explainable World-Level Reasoning.}}
Post-training internalizes evidence selection and reasoning style so that inference depends less on external controllers.
Within Layer~4, reinforcement learning and preference alignment encode how evidence should be gathered and how explanations should be structured, as explored by VidGuard-R1, Veritas, and BusterX++~\citep{Park2025VidGuardR1,Tan2025Veritas,Wen2025BusterXpp}.
DeeptraceReward learns complementary reward supervision aligned to human-perceived fakeness in AI-generated videos~\citep{Fu2025DeeptraceReward}, and VideoVeritas further couples preference alignment with perception pretext reinforcement learning and fact-based reasoning for more balanced AIGC-V verification~\citep{tan2026videoveritas}.
Related efforts extend post-training to multi-stage reasoning and modular explanation under complex AIGC-V settings~\citep{wen2025busterx,sun2025edvd,li2025skyraaigeneratedvideodetection,chen2024x2dfd}. The emphasis shifts from generic instruction-following to disciplined justification, including when to retrieve, when to localize uncertainty, and how to present compact but auditable evidence.

\begin{surveytakeaways}[Takeaways for Layer 4]
  \item \textbf{Why It Works:} It targets a harder bottleneck than within-video agreement: a clip may look self-consistent yet still fail once its implied claims are checked against facts, commonsense, causality, or physical rules.
  \item \textbf{Paradigm Fit:} Best for GVS and claim-centric LMV or AVE cases where the decision turns on external verification of actors, events, relations, or physical plausibility, especially when within-video modalities appear mutually consistent.
  \item \textbf{Failure Modes:} Verification depends on knowledge coverage, retrieval freshness, and correct claim decomposition, and VLMs can hallucinate, over-abstract, or miss subtle visual evidence while sounding confident.
\end{surveytakeaways}

\begin{table}[!htbp]
\centering
\fontsize{7.1}{8.15}\selectfont
{\setlength{\tabcolsep}{1.8pt}
\renewcommand{\arraystretch}{1.2}
\begin{tabularx}{\linewidth}{@{}>{\raggedright\arraybackslash}p{0.22\linewidth} Y Y Y c@{}}
\toprule
\textbf{Method} & \textbf{Base Model and System} & \textbf{Training} & \textbf{Output} & \textbf{Date} \\
\midrule
\rowcolor{metablue!12}
\multicolumn{5}{@{}l}{\textbf{\textcolor{metablue!90!black}{A. Prompts and adapters for representation calibration}}} \\
\addlinespace[0.08em]
\textbf{ChatGPT Detect}~\citep{Jia2024CVPRW,Shahzad2025ChatGPTAV}
& ChatGPT
& Prompt
& Verdict+exp.
& 06/2024 \\

\textbf{CPML}~\citep{Lai2024CPML,DAmelio2023rPPGNote}
& rPPG+landmark encoder
& Prompt-guided
& Score
& 11/2024 \\

\textbf{DeepFake-Adapter}~\citep{shao2024deepfakeadapterdualleveladapterdeepfake}
& ViT (frozen)
& Adapter tuning
& Frame label (agg.)
& 01/2025 \\

\textbf{RepDFD}~\citep{Lin2025RepDFD}
& Frozen VLM
& Prompt tuning
& Label
& 04/2025 \\

\textbf{AuthGuard}~\citep{Shen2025AuthGuard}
& Vision encoder + LLM
& Cls+ITC; uncertainty
& Verdict+exp.
& 06/2025 \\

\textbf{LVLMDFD}~\citep{Yu2025LVLMDFD}
& LVLM + LLM
& Detector+prompt learner
& Verdict+loc.+exp.
& 07/2025 \\

\addlinespace[0.12em]
\cdashline{1-5}[2pt/2pt]
\addlinespace[0.12em]
\rowcolor{metablue!12}
\multicolumn{5}{@{}l}{\textbf{\textcolor{metablue!90!black}{B. Tool-augmented agents for evidence gathering}}} \\
\addlinespace[0.08em]
\textbf{LAVID}~\citep{Liu2025LAVID}
& LVLM agent + tools
& Tool loop
& Verdict+evidence
& 02/2025 \\

\textbf{DAVID-XR1}~\citep{gao2025david}
& Video VLM
& SFT
& Loc.+defect reasoning
& 06/2025 \\

\textbf{FakeHunter}~\citep{chen2025memory}
& Vision+audio encoders + VLM agent
& Agent+retrieval
& Verdict+evidence
& 08/2025 \\

\textbf{DeepAgent}~\citep{zaman2025deepagent}
& Multi-agent pipeline
& Not stated
& MM verdict
& 12/2025 \\

\addlinespace[0.12em]
\cdashline{1-5}[2pt/2pt]
\addlinespace[0.12em]
\rowcolor{metablue!12}
\multicolumn{5}{@{}l}{\textbf{\textcolor{metablue!90!black}{C. Post-training, preferences and rewards}}} \\
\addlinespace[0.08em]
\textbf{X2-DFD}~\citep{chen2024x2dfd}
& LLaVA + aux detectors
& SFT (explainable data)
& Verdict+exp.
& 10/2024 \\

\textbf{BusterX}~\citep{wen2025busterx}
& Qwen2.5-VL
& SFT $\rightarrow$ RL
& Verdict+exp.
& 05/2025 \\

\textbf{BusterX++}~\citep{Wen2025BusterXpp}
& Qwen2.5-VL
& RL $\rightarrow$ SFT $\rightarrow$ RL
& Verdict+struct. exp.
& 07/2025 \\

\textbf{VERITAS}~\citep{Tan2025Veritas}
& InternVL3 and Qwen2.5-VL
& SFT $\rightarrow$ MiPO $\rightarrow$ P-GRPO
& Verdict+exp.
& 08/2025 \\

\textbf{DeeptraceReward}~\citep{Fu2025DeeptraceReward}
& VideoLLaMA3 and Qwen2.5-VL with reward modeling
& Reward-model train
& Reward
& 09/2025 \\

\textbf{VidGuard-R1}~\citep{Park2025VidGuardR1}
& Qwen2.5-VL
& SFT $\rightarrow$ DPO $\rightarrow$ GRPO
& Verdict+exp.
& 10/2025 \\

\textbf{EDVD-LLaMA}~\citep{sun2025edvd}
& Qwen2.5-7B + video encoder
& SFT
& Verdict+exp.
& 10/2025 \\

\textbf{Skyra and Skyra-RL}~\citep{li2025skyraaigeneratedvideodetection}
& Qwen2.5-VL
& SFT $\rightarrow$ RL
& Verdict+grounded reasoning
& 12/2025 \\

\textbf{VideoVeritas}~\citep{tan2026videoveritas}
& Video MLLM
& Pref. align + PPRL
& Verdict+reasoning
& 02/2026 \\
\bottomrule
\end{tabularx}}
\caption{\textbf{Layer 4: Language-guided world-level reasoning.} Representative methods grouped by (A) prompts and adapters, (B) tool-augmented agents, and (C) post-training with preferences and rewards. \textit{Base Model and System} reports backbone names when available.}
\label{tab:l4_summary}
\end{table}

\FloatBarrier

\subsection{Quantitative Summary of the Four-Layer Landscape}
\label{sec:quantitative-summary}

\subsubsection{Year-wise Layer Distribution}
\label{sec:yearwise-layer-distribution}
Table~\ref{tab:yearly-layer-share} summarizes the surveyed detection papers discussed in Section~\ref{sec:methods} and tracks how the four-layer taxonomy evolves over time.
We report both yearly share and cumulative share. The yearly ratio reflects release-time dynamics within each publication year, whereas the cumulative ratio shows how the full surveyed landscape shifts as methods accumulate.
Because 2026 is still in progress, the table reports complete publication years through 2025; the cumulative denominator additionally includes the three surveyed papers before 2020.
Two patterns stand out. First, the early landscape is still overwhelmingly visual-view: from 2020 to 2022, Layers~1 and~2 account for 30 of 37 surveyed methods, showing that the field was primarily organized around forensic traces and temporal coherence. Second, the transition toward the language view is gradual rather than monotonic. The yearly language-view share rises from 7.7\% in 2020 to 26.7\% and 22.2\% in 2021--2022, reaches 40.0\% in 2023, and then expands further in 2024 and 2025.
The decisive break arrives in 2025, when 30 of 48 surveyed methods fall into the language view. The cumulative language-view share correspondingly grows from 6.3\% to 40.3\%, indicating that multimodal verification and world-level reasoning have moved from a marginal line of work to a major axis of recent AIGC-V detection. At the same time, the visual view remains substantial, with 18 methods still belonging to Layers~1 and~2 in 2025; low-level and temporal evidence are therefore not replaced, but increasingly complemented by higher-level verification.

\providecommand{\yearviscountshade}[2]{\textcolor{#1}{\textbf{#2}}}
\providecommand{\yearlangcountshade}[2]{\textcolor{#1}{\textbf{#2}}}
\providecommand{\yearlangsharecell}[3]{\makecell[c]{\textcolor{#1}{\textbf{#2}}\\{\scriptsize #3}}}

\begin{table}[H]
  \centering
  \footnotesize
  \setlength{\tabcolsep}{2.5pt}
  \renewcommand{\arraystretch}{1.12}
  \begin{tabular}{@{}C{0.06\linewidth}*{4}{C{0.055\linewidth}}C{0.065\linewidth}C{0.07\linewidth}C{0.095\linewidth}C{0.16\linewidth}C{0.16\linewidth}@{}}
    \toprule
    \multirow{2}{*}{\textbf{Year}}
    & \multicolumn{5}{c}{\textbf{Layer Counts}}
    & \multicolumn{2}{c}{\textbf{View Counts}}
    & \multicolumn{2}{c}{\textbf{Language-View Share}} \\
    \cmidrule(lr){2-6}\cmidrule(lr){7-8}\cmidrule(lr){9-10}
    & \textbf{L1} & \textbf{L2} & \textbf{L3} & \textbf{L4} & \textbf{Total}
    & \textcolor{morandiblue!82!black}{\textbf{Visual}} & \textcolor{morandired!76!black}{\textbf{Language}}
    & \makecell[c]{\textbf{Yearly}}
    & \makecell[c]{\textbf{Cumulative}} \\
    \midrule
    \textbf{2020} & 5 & 7 & 1 & 0 & 13 & \yearviscountshade{morandiblue!78!white}{12} & \yearlangcountshade{morandired!40!white}{1}
    & \yearlangsharecell{morandired!42!white}{7.7\%}{1 of 13}
    & \yearlangsharecell{morandired!40!white}{6.3\%}{1 of 16} \\
    \textbf{2021} & 5 & 6 & 4 & 0 & 15 & \yearviscountshade{morandiblue!74!white}{11} & \yearlangcountshade{morandired!52!white}{4}
    & \yearlangsharecell{morandired!52!white}{26.7\%}{4 of 15}
    & \yearlangsharecell{morandired!46!white}{16.1\%}{5 of 31} \\
    \textbf{2022} & 4 & 3 & 2 & 0 & 9 & \yearviscountshade{morandiblue!60!white}{7} & \yearlangcountshade{morandired!46!white}{2}
    & \yearlangsharecell{morandired!46!white}{22.2\%}{2 of 9}
    & \yearlangsharecell{morandired!44!white}{17.5\%}{7 of 40} \\
    \textbf{2023} & 3 & 3 & 4 & 0 & 10 & \yearviscountshade{morandiblue!56!white}{6} & \yearlangcountshade{morandired!52!white}{4}
    & \yearlangsharecell{morandired!60!white}{40.0\%}{4 of 10}
    & \yearlangsharecell{morandired!50!white}{22.0\%}{11 of 50} \\
    \textbf{2024} & 9 & 5 & 4 & 3 & 21 & \yearviscountshade{morandiblue!80!white}{14} & \yearlangcountshade{morandired!64!white}{7}
    & \yearlangsharecell{morandired!64!white}{33.3\%}{7 of 21}
    & \yearlangsharecell{morandired!54!white}{25.4\%}{18 of 71} \\
    \textbf{2025} & 0 & 18 & 15 & 15 & 48 & \yearviscountshade{morandiblue!80!white}{18} & \yearlangcountshade{morandired!88!white}{30}
    & \yearlangsharecell{morandired!88!white}{62.5\%}{30 of 48}
    & \yearlangsharecell{morandired!66!white}{40.3\%}{48 of 119} \\
    \midrule
    \rowcolor{morandiblue!8}
    \textit{$\Delta$} & \textit{-5} & \textit{+11} & \textit{+14} & \textit{+15} & \textit{+35} & \textit{\textcolor{morandiblue!58!white}{+6}} & \textit{\textcolor{morandired!80!white}{+29}} & \textit{\textcolor{morandired!88!white}{+54.8}} & \textit{\textcolor{morandired!70!white}{+34.0}} \\
    \bottomrule
  \end{tabular}
  \caption{Year-wise quantitative summary for the surveyed detection papers discussed in Section~\ref{sec:methods}. The visual view combines Layers~1 and~2, whereas the language view combines Layers~3 and~4. Share values are reported as a percentage with the corresponding count ratio. The table reports complete publication years through 2025; the cumulative denominator additionally includes the three surveyed papers before 2020. In the $\Delta$ row, differences are computed between 2020 and 2025, and the two share columns report percentage-point changes.}
  \label{tab:yearly-layer-share}
\end{table}

\subsubsection{Layer-wise Performance Snapshot}
\label{sec:layer-performance-snapshot}
To complement the taxonomy with a compact quantitative anchor, Table~\ref{tab:layer-performance-snapshot} collects reported AUC (\%) for representative methods across the four layers.
Because reporting protocols remain heterogeneous, each number is paired with its protocol label, either cross-dataset (CD) or in-domain (ID). The table should therefore be read as a protocol-aware snapshot rather than as a single cross-layer leaderboard.
Three observations are useful when reading the table. First, visual-view methods remain highly competitive under standard CD evaluation: strong Layer~1 and Layer~2 systems still achieve high AUC on CDFv2 and retain reasonable transfer to DFDC, which shows that low-level and temporal evidence continue to generalize well when benchmarks emphasize generator shift. Second, Layer~4 methods are already competitive under the same CD setup. RepDFD and LVLMDFD match or exceed lower-layer baselines on CDFv2 and DFDCP, suggesting that language-guided verification can be added without giving up benchmark strength. Third, the table also reveals an evaluation gap for higher layers: Layer~3 is represented here only by an ID result, and many recent multimodal or world-grounded systems still lack standardized CD reporting.
This means the table should not be read as evidence that later layers uniformly outperform earlier ones. Rather, it shows that different layers answer different verification needs, while current benchmark practice is still better aligned with artifact and coherence detection than with claim verification or world-level reasoning. Across the CD rows, DFDC also remains the hardest benchmark, with even strong methods mostly landing in the mid-70s to low-80s, which further cautions against naive rank ordering.

\begin{table}[H]
  \centering
  \footnotesize
  \setlength{\tabcolsep}{2pt}
  \renewcommand{\arraystretch}{1.04}%
\begin{tabularx}{\linewidth}{@{}L{0.37\linewidth}C{0.07\linewidth}C{0.09\linewidth}C{0.13\linewidth}C{0.13\linewidth}C{0.13\linewidth}@{}}
    \toprule
    \textbf{Method} & \textbf{Layer} & \textbf{Protocol} & \textbf{CDFv2} & \textbf{DFDCP} & \textbf{DFDC} \\
    \midrule
    \multicolumn{6}{@{}l}{\textbf{\textcolor{metablue!90!black}{Layer 1}}} \\
    \textbf{FreqBlender}~\citep{zhou2024freqblender} & L1 & CD & 94.6 & 87.6 & 74.6 \\
    \textbf{SeeABLE}~\citep{larue2023seeable} & L1 & CD & 87.3 & 86.3 & 75.9 \\
    \textbf{LSDA}~\citep{DBLP:conf/cvpr/YanLLLW24} & L1 & CD & 83.0 & 81.5 & 73.6 \\
    \textbf{Style Latent Flows}~\citep{choi2024exploiting} & L1 & CD & 89.0 & -- & -- \\
    \addlinespace[0.25em]
    \cdashline{1-6}[2pt/2pt]
    \addlinespace[0.20em]
    \multicolumn{6}{@{}l}{\textbf{\textcolor{metablue!90!black}{Layer 2}}} \\
    \textbf{TALL-Swin}~\citep{xu2023tall} & L2 & CD & 90.8 & -- & 76.8 \\
    \textbf{LipForensics}~\citep{haliassos2021lips} & L2 & CD & 82.4 & -- & 73.5 \\
    \textbf{Two-branch}~\citep{masi2020two} & L2 & CD & 76.6 & -- & -- \\
    \addlinespace[0.25em]
    \cdashline{1-6}[2pt/2pt]
    \addlinespace[0.20em]
    \multicolumn{6}{@{}l}{\textbf{\textcolor{metablue!90!black}{Layer 3}}} \\
    \textbf{MDS}~\citep{1a23bed3dbff4e9cb0984b74aff3376a} & L3 & ID & -- & -- & 90.6 \\
    \addlinespace[0.25em]
    \cdashline{1-6}[2pt/2pt]
    \addlinespace[0.20em]
    \multicolumn{6}{@{}l}{\textbf{\textcolor{metablue!90!black}{Layer 4}}} \\
    \textbf{RepDFD}~\citep{Lin2025RepDFD} & L4 & CD & 89.9 & 95.0 & 81.0 \\
    \textbf{LVLMDFD}~\citep{Yu2025LVLMDFD} & L4 & CD & 94.3 & 92.4 & 77.0 \\
    \bottomrule
  \end{tabularx}
  \caption{AUC (\%) snapshot across the four-layer taxonomy with explicit protocol labels. CD = cross-dataset with training on FF++ and testing on CDFv2, DFDCP, and DFDC; ID = in-domain. CD and ID values are \textbf{not directly comparable}, so the table should be read as a protocol-aware snapshot rather than a single ranked summary. ``--'' denotes not reported.}
  \label{tab:layer-performance-snapshot}
\end{table}

\FloatBarrier

\section{Evaluation of AIGC-V detection}
\label{sec:evaluation}
We provide a comprehensive overview of a generic evaluation framework for AIGC-V detection, focusing on evaluation metrics under the dual-view setting and benchmarks organized in line with three AIGC-V paradigms.
We also summarize adjacent non-detector-first diagnostic resources for synthetic-video factual-fidelity evaluation.

\subsection{Evaluation Metrics}
\label{sec:eval-metrics}
\paragraph{\textit{\textbf{Shared Basic Metrics.}}}
We report standard binary metrics for real-versus-AIGC-V detection, including \textit{Acc}, \textit{AUC}, \textit{Precision}, \textit{Recall}, \textit{F1}, \textit{EER}, and \textit{PR-AUC} under class imbalance.
Scores may be computed at the frame level or video level, with temporal aggregation such as mean pooling or voting.
These metrics provide a shared baseline, but they do not by themselves diagnose temporal coherence, physical plausibility, or semantic consistency.

\paragraph{\textit{\textbf{Visual View Metrics.}}}
The visual view evaluates whether a detector captures perceptual evidence that a video \emph{looks and moves} like a real capture.
Two aspects matter.
\textbf{(i) Intrinsic-cue robustness:} acquisition and generation artifacts such as sensor and ISP traces, resampling and coding footprints, and sampling artifacts are still commonly measured with \textit{Acc}, \textit{AUC}, and \textit{EER}~\citep{cheng2024can,zhou2024freqblender,gu2022hierarchical}, but the decisive protocols are cross-dataset transfer and perturbation stress tests over codec, bitrate, and resolution. Fixed-operating-point metrics such as \textit{TPR@FPR=$\alpha$} are also important in deployment-oriented settings.
\textbf{(ii) Spatiotemporal and physical consistency:} evaluation asks whether motion, layout, and interactions remain temporally and physically plausible~\citep{feng2023self,zhang2024learning,anand2025detecting,zhang2025physics,zheng2025d3,kim2025beyond,xu2024learning}. Video-level reporting, temporal-perturbation drops, and motion-ablation studies are generally more informative than strong frame-level scores alone~\citep{kundu2025towards,nie2024dip,yan2025generalizing}.
Overall, visual-view evaluation should prioritize video-level robustness rather than closed-set frame discrimination.

\paragraph{\textit{\textbf{Language View Metrics.}}}
The language view evaluates whether video, audio, and accompanying text or metadata jointly support plausible \emph{facts about the world}.
It also decomposes into two aspects.
\textbf{(i) Cross-modal alignment:} benchmarks probe lip-audio synchrony, speaker identity consistency, caption-video matching, and temporally localized mismatches. Depending on the setting, works combine \textit{Acc} and \textit{AUC} with \textit{AP}, \textit{AR}, \textit{Recall@K}, and \textit{mAP} over \textit{IoU} thresholds, often under modality corruption or degraded-channel tests~\citep{katamneni2024contextualcrossmodalattentionaudiovisual,Xu_2025,wang2025fauforensicsboostingaudiovisualdeepfake,anshul2025next}.
\textbf{(ii) World knowledge and reasoning:} for fact-level and narrative-level verification, standard classification scores are insufficient. Evaluation therefore adds human judgments, pairwise preferences, question answering, and rationale-quality metrics such as \textit{BLEU}, \textit{ROUGE-L}, \textit{METEOR}, \textit{CIDEr}, and embedding-based similarity~\citep{Yu2025LVLMDFD,Shen2025AuthGuard,Wen2025BusterXpp,sun2025edvd,gao2025david,Fu2025DeeptraceReward,hondru2025exddv}. Recent studies further show that observers rely on mixed visual, vocal, and knowledge cues, and that detector outputs can mislead downstream claim verification when they are not tied to explicit evidence~\citep{chen2026seeing,shi2026watch,sagar2026fact}.
In short, language-view evaluation shifts the question from ``does it look real'' to ``does it support a plausible, well-grounded story.''

\begin{surveytakeaways}[What Evaluation Must Show]
  \item \textbf{AUC is necessary but insufficient:} it measures separability on a protocol, not whether the decision is evidentially well grounded.
  \item \textbf{Protocols should follow the claimed evidence pathway:} robustness under shift for visual cues, and alignment, localization, grounding, and claim verification with reasoning.
  \item \textbf{Benchmark numbers are not interchangeable.} In-domain accuracy, cross-dataset transfer, operating-point robustness, and explanation quality answer different questions and should not collapse into one headline score.
\end{surveytakeaways}
\vspace{-0.35em}

\subsection{Benchmarks}
\label{sec:benchmarks}
Benchmark development still follows the three AIGC-V paradigms in Section~\ref{sec:paradigms}, but the three branches now differ not only in scale, but also in annotation granularity and evidential reach. Taken together, these three paradigm-specific benchmark families define the core detector-evaluation landscape. 
Table~\ref{tab:benchmarks} summarizes this landscape of benchmarks for AIGC-V detection. A complementary line of adjacent diagnostic resources, discussed later in this subsection, broadens that landscape beyond detector-first evaluation toward physical-rule testing, world-dynamics and causality probes, and explanation-oriented diagnosis.

\paragraph{\textbf{LMV Related Benchmarks.}}
LMV related benchmarks remain the historical backbone of AIGC-V detection because partial edits preserve most of the source video while exposing localized forensic traces. That structure makes LMV especially suitable for testing artifact sensitivity, compression robustness, codec variation, and cross-dataset transfer. Among the foundational resources, FaceForensics++~\citep{rossler2019faceforensics++} was decisive because it standardized post-compression manipulation protocols and effectively set the early evaluation culture; Celeb-DF~\citep{li2020celeb} and DFDC~\citep{DFDC2020} then pushed the field toward harder realism and larger scale; DeeperForensics-1.0~\citep{Jiang2020DeeperForensics-1.0} made robustness under real-world distortions a central concern rather than an afterthought. ForgeryNet~\citep{He_2021_CVPR} broadened the manipulation space further. Later resources pushed LMV in more deployment-facing directions: WildDeepfake~\citep{zi20WildDeepfake} and KoDF~\citep{kwon2021kodf} moved toward in-the-wild and cross-lingual conditions, AI-Face~\citep{lin2025ai} enlarged demographic coverage, and CDDB~\citep{Li_2023_WACV} together with DeepfakeBench~\citep{DeepfakeBench_YAN_NEURIPS2023} reframed LMV as a protocol problem rather than a single-dataset problem. DD-VQA~\citep{zhang2024common} and ExDDV~\citep{hondru2025exddv} begin to attach question answering and explanation supervision to a regime that was originally almost purely artifact-driven, while Beyond Static Artifacts~\citep{gu2026beyond} turns temporal deepfake reasoning itself into an explicit benchmark target for VLMs.

\paragraph{\textbf{AVE Related Benchmarks.}}
AVE related benchmarks are organized around whether speech, mouth motion, speaker identity, and language-bearing cues remain jointly coherent over time, so they are less about static artifact capture than about temporal and cross-modal verification. This branch remains smaller than LMV partly because realistic audio-visual edits require synchronized manipulation across modalities together with finer temporal annotation. FakeAVCeleb~\citep{FakeAVCeleb2021} established the basic paired audio-video manipulation setting, but LAV-DF~\citep{cai2022you} was especially important because it turned localized, content-driven audio-visual mismatch into a first-class benchmark target rather than treating the clip as uniformly fake. Scale then increased with FakeMix~\citep{jung2024and} and AV-Deepfake1M~\citep{cai2024av}. The more recent resources differentiate the evaluation question itself: DigiFakeAV~\citep{liu2025beyond} targets diffusion-based digital humans, VCapAV~\citep{wang2025vcapav} matters because it converts AVE into a caption-grounded inconsistency problem, MAVOS-DD~\citep{croitoru2025mavos} studies multilingual open-set conditions, X-AVFake~\citep{chen2025memory} introduces grounded language reasoning, and MMDF~\citep{kim2026xavdt} broadens manipulation and generator coverage. Evaluation in this branch therefore centers on lip-audio alignment, speaker-content consistency, dubbed or spliced segments, temporal localization, and robustness under degraded settings.

\paragraph{\textbf{GVS Related Benchmarks.}}
GVS related benchmarks are the fastest-moving branch because fully generated videos weaken explicit editing traces while amplifying generator diversity, plausibility failures, and transfer risk. Early resources, including GVF~\citep{ma2024detecting}, GenVidDet~\citep{ji2024distinguish}, GenVideo~\citep{DeMamba}, DVF~\citep{NEURIPS2024_dccbeb7a}, and GenVidBench~\citep{ni2025genvidbench}, established fully generated video detection as a distinct benchmark regime rather than an extension of face-forgery evaluation. Later benchmarks broadened both scope and supervision: LOKI~\citep{ye2024loki} and Deepfake-Eval-2024~\citep{chandra2025deepfake} expanded modality coverage and in-the-wild conditions; GenWorld~\citep{chen2025genworlddetectingaigeneratedrealworld} shifted attention toward real-world simulation; DAVID-X~\citep{gao2025david} and DeeptraceReward~\citep{Fu2025DeeptraceReward} introduced defect-level or human-perceived trace supervision; and ER-FF++set~\citep{sun2025edvd} together with AEGIS~\citep{10.1145/3746027.3758295} continued the move toward explanation-aware evaluation. The newest large-scale suites, including AIGVDBench~\citep{DBLP:journals/corr/abs-2601-11035}, SynthForensics~\citep{leotta2026synthforensics}, ViFBench~\citep{li2025skyraaigeneratedvideodetection}, Video Reality Test~\citep{wang2025videorealitytestaigenerated}, and MintVid~\citep{tan2026videoveritas}, emphasize generator diversity, stronger realism, and deployment-oriented stress testing. AIGVDBench is particularly consequential because its scale makes cross-generator transfer patterns interpretable at ecosystem level. As a result, GVS benchmarks increasingly evaluate cross-generator transfer, physical plausibility, event logic, and semantically fabricated scenarios rather than artifact sensitivity alone.

\begin{surveytakeaways}[What Benchmarks Actually Cover]
  \item \textbf{Coverage is structurally uneven.} LMV has the deepest protocol history, AVE remains narrower because synchronized multimodal annotation is costly, and GVS changes fastest as generator turnover keeps resetting the task.
  \item \textbf{The three paradigms test different evidence pathways.} LMV emphasizes localized forensic residue and transfer, AVE temporal and cross-modal coherence, and GVS transfer under weaker artifacts together with plausibility and event logic.
  \item \textbf{Benchmark difficulty shifts with the generation paradigm.} LMV mainly asks whether traces survive compression and domain shift, AVE whether inconsistencies can be localized across modalities, and GVS whether detectors remain reliable as generators become more realistic and diverse.
\end{surveytakeaways}

\begin{table}[p]
\centering
\scriptsize
{\setlength{\tabcolsep}{1pt}
\renewcommand{\arraystretch}{1.0}
\begin{tabularx}{\linewidth}{@{}L{0.28\linewidth}L{0.52\linewidth}C{0.10\linewidth}C{0.07\linewidth}@{}}
\toprule
\multicolumn{1}{@{}L{0.28\linewidth}}{\textbf{\textcolor{metablue!90!black}{Benchmark \& Dataset}}}
& \multicolumn{1}{L{0.52\linewidth}}{\textbf{\textcolor{metablue!90!black}{Description}}}
& \multicolumn{1}{C{0.10\linewidth}}{\textbf{\textcolor{metablue!90!black}{Paradigms}}}
& \multicolumn{1}{C{0.07\linewidth}@{}}{\textbf{\textcolor{metablue!90!black}{Date}}} \\
\midrule
\multicolumn{4}{@{}l}{\textbf{\textcolor{metablue!90!black}{A. Local Manipulation Video (LMV)}}} \\
\addlinespace[0.14em]
\textbf{FaceForensics++}~\citep{rossler2019faceforensics++} & Forensics dataset with 1,000 original videos. & \textit{LMV} & 01/2019 \\
\textbf{Celeb-DF}~\citep{li2020celeb} & Large-scale challenging deepfake-forensics dataset. & \textit{LMV} & 09/2019 \\
\textbf{DeeperForensics-1.0}~\citep{Jiang2020DeeperForensics-1.0} & Large-scale dataset for real-world face-forgery detection. & \textit{LMV} & 05/2020 \\
\textbf{DFDC}~\citep{DFDC2020} & Large-scale face-swapping dataset (mostly local forgeries). & \textit{LMV} & 06/2020 \\
\textbf{WildDeepfake}~\citep{zi20WildDeepfake} & In-the-wild dataset for AIGC-V detection. & \textit{LMV} & 10/2020 \\
\textbf{ForgeryNet}~\citep{He_2021_CVPR} & Mega-scale benchmark for image and video face-forgery analysis. & \textit{LMV} & 06/2021 \\
\textbf{KoDF}~\citep{kwon2021kodf} & Large-scale Korean AIGC-V detection dataset. & \textit{LMV} & 10/2021 \\
\textbf{CDDB}~\citep{Li_2023_WACV} & Benchmark for easy, hard, and long-sequence AIGC-V detection. & \textit{LMV} & 01/2023 \\
\textbf{DF-Platter}~\citep{Narayan_2023_CVPR} & Multi-face heterogeneous deepfake dataset. & \textit{LMV} & 06/2023 \\
\textbf{DeepfakeBench}~\citep{DeepfakeBench_YAN_NEURIPS2023} & Comprehensive AIGC-V detection benchmark. & \textit{LMV} & 07/2023 \\
\textbf{AI-Face}~\citep{lin2025ai} & Million-scale, demographically annotated AI-face dataset. & \textit{LMV\&GVS} & 06/2024 \\
\textbf{DD-VQA}~\citep{zhang2024common} & AIGC-V detection VQA (image--question--answer triplets). & \textit{LMV} & 09/2024 \\
\textbf{ExDDV}~\citep{hondru2025exddv} & Benchmark for explainable AIGC-V video detection. & \textit{LMV} & 11/2025 \\
\textbf{FAQ / Beyond Static Artifacts}~\citep{gu2026beyond} & Forensic benchmark and instruction-tuning resource for temporal video deepfake reasoning in VLMs. & \textit{LMV} & 02/2026 \\

\addlinespace[0.10em]
\cdashline{1-4}[2pt/2pt]
\addlinespace[0.06em]
\multicolumn{4}{@{}l}{\textbf{\textcolor{metablue!90!black}{B. Audio-Visual Editing (AVE)}}} \\
\addlinespace[0.14em]
\textbf{FakeAVCeleb}~\citep{FakeAVCeleb2021} & Audio-visual multimodal deepfake dataset. & \textit{AVE} & 08/2021 \\
\textbf{LAV-DF}~\citep{cai2022you} & Content-driven localized audio-visual deepfake dataset. & \textit{AVE} & 11/2022 \\
\textbf{FakeMix}~\citep{jung2024and} & Clip-level benchmark for manipulated audio and video segments. & \textit{AVE} & 08/2024 \\
\textbf{AV-Deepfake1M}~\citep{cai2024av} & Large-scale LLM-driven audio-visual deepfake dataset. & \textit{AVE} & 10/2024 \\
\textbf{ArEnAV}~\citep{kuckreja2025tell} & Audio-visual deepfake dataset for Arabic--English code-switching (CSW). & \textit{AVE} & 05/2025 \\
\textbf{MAVOS-DD}~\citep{croitoru2025mavos} & Multilingual open-set benchmark for audio-visual AIGC-V detection. & \textit{AVE} & 05/2025 \\
\textbf{DigiFakeAV}~\citep{liu2025beyond} & Benchmark for diffusion-based digital-human audio-visual forgeries. & \textit{AVE} & 05/2025 \\
\textbf{SocialDF}~\citep{batra2025socialdf} & 2,126 social-media videos with real and deepfake content under SOTA manipulations. & \textit{AVE} & 07/2025 \\
\textbf{VCapAV}~\citep{wang2025vcapav} & Video-caption-based audio-visual AIGC-V detection dataset. & \textit{AVE} & 08/2025 \\
\textbf{X-AVFake}~\citep{chen2025memory} & Dual-modality manipulations with grounded language reasoning. & \textit{AVE} & 08/2025 \\
\textbf{AV-Deepfake1M++}~\citep{cai2025av} & 2M clips extending AV-Deepfake1M with diverse audio-visual manipulations. & \textit{AVE} & 10/2025 \\
\textbf{MMDF}~\citep{kim2026xavdt} & Multimodal deepfake dataset spanning GAN, diffusion, and flow-matching manipulations. & \textit{AVE} & 03/2026 \\

\addlinespace[0.10em]
\cdashline{1-4}[2pt/2pt]
\addlinespace[0.06em]
\multicolumn{4}{@{}l}{\textbf{\textcolor{metablue!90!black}{C. Generative Video Synthesis (GVS)}}} \\
\addlinespace[0.14em]
\textbf{GVF}~\citep{ma2024detecting} & Generated Video Forensics benchmark for AI-video detection. & \textit{GVS} & 02/2024 \\
\textbf{GenVidDet}~\citep{ji2024distinguish} & Real and AIGC-V videos from eight generation models. & \textit{GVS} & 05/2024 \\
\textbf{GenVideo}~\citep{DeMamba} & AIGC-V detection dataset. & \textit{GVS} & 05/2024 \\
\textbf{DVF}~\citep{NEURIPS2024_dccbeb7a} & Diffusion Video Forensics dataset and benchmark. & \textit{GVS} & 12/2024 \\
\textbf{GenVidBench}~\citep{ni2025genvidbench} & AIGC-V detection dataset. & \textit{GVS} & 01/2025 \\
\textbf{Deepfake-Eval-2024}~\citep{chandra2025deepfake} & Multi-modal in-the-wild benchmark of deepfakes from 2024. & \textit{L\&A\&G} & 03/2025 \\
\textbf{LOKI}~\citep{ye2024loki} & Multimodal synthetic-data detection benchmark over video, image, 3D, text, and audio. & \textit{GVS} & 04/2025 \\
\textbf{GenBuster-200K}~\citep{wen2025busterx} & Real + synthetic videos simulating real-world conditions. & \textit{GVS} & 05/2025 \\
\textbf{GenWorld}~\citep{chen2025genworlddetectingaigeneratedrealworld} & Large-scale real-world simulation dataset for AI-video detection. & \textit{GVS} & 06/2025 \\
\textbf{Ivy-Fake}~\citep{jiang2025ivy} & Large-scale benchmark for explainable AIGC detection. & \textit{GVS} & 06/2025 \\
\textbf{DAVID-X}~\citep{gao2025david} & AIGC-V with defect-level spatiotemporal annotations + rationales. & \textit{GVS} & 06/2025 \\
\textbf{GenBuster++}~\citep{Wen2025BusterXpp} & A cross-modal benchmark for VLM evaluation. & \textit{GVS} & 07/2025 \\
\textbf{DeeptraceReward}~\citep{Fu2025DeeptraceReward} & Spatiotemporal benchmark with human-perceived fake traces. & \textit{GVS} & 09/2025 \\
\textbf{ER-FF++set}~\citep{sun2025edvd} & Benchmark with supervision for detection + explanation. & \textit{GVS} & 10/2025 \\
\textbf{AEGIS}~\citep{10.1145/3746027.3758295} & Large-scale benchmark for AIGC-V authenticity. & \textit{GVS} & 10/2025 \\
\textbf{ViFBench}~\citep{li2025skyraaigeneratedvideodetection} & 3K videos from 10+ SOTA generators. & \textit{GVS} & 12/2025 \\
\textbf{Video Reality Test}~\citep{wang2025videorealitytestaigenerated} & An ASMR-sourced benchmark. & \textit{GVS} & 12/2025 \\
\textbf{AIGVDBench}~\citep{DBLP:journals/corr/abs-2601-11035} & Large-scale benchmark with 440K videos from 31 generation models and 33 evaluated detectors. & \textit{GVS} & 01/2026 \\
\textbf{SynthForensics}~\citep{leotta2026synthforensics} & Human-centric synthetic-video benchmark with 6,815 videos from five open-source T2V generators. & \textit{GVS} & 02/2026 \\
\textbf{MintVid}~\citep{tan2026videoveritas} & Lightweight high-quality benchmark with 3K videos from nine state-of-the-art generators. & \textit{GVS} & 02/2026 \\
\bottomrule
\end{tabularx}}
\caption{Overview of existing datasets and benchmarks through March 2026, and their alignment with the AIGC-V paradigms introduced in our survey.}
\label{tab:benchmarks}
\end{table}

\paragraph{\textbf{Diagnostic Resources Beyond Detection Benchmarks.}}
Adjacent diagnostic resources extend evaluation along a clear progression: from basic rule violations, to coherent world dynamics, and finally to explanation-oriented diagnosis. Physical Rule Violations asks whether generated videos obey basic physical constraints: VideoPhy~\citep{bansal2025videophy}, Physics-IQ~\citep{motamed2025generative}, and IPV-Bench~\citep{bai2025impossiblevideos} stress physical commonsense and impossible scenarios; Morpheus~\citep{zhang2025morpheus} and T2VPhysBench~\citep{guo2025t2vphysbench} tighten this into experiment-grounded and first-principles testing; and PhyWorldBench~\citep{gu2025phyworldbench}, VideoPhy-2~\citep{bansal2026videophy2}, and Physion-Eval~\citep{zhang2026physioneval} move further toward action-centric settings, localized glitches, and expert reasoning traces. World Dynamics and Causality then asks whether those local regularities scale to temporally coherent events and world knowledge over time. WorldSimBench~\citep{qin2025worldsimbench} makes the world-simulator question explicit; PhyGenBench~\citep{meng2025towardsworldsimulator} studies multi-step intuitive physics; StoryEval~\citep{wang2024storyeval} turns consecutive events into a story-level stress test; VideoVerse~\citep{wang2025videoverse} and T2VWorldBench~\citep{chen2025t2vworldbench} probe temporal causality, hidden semantics, and world knowledge; and SVBench~\citep{peng2025svbench} extends the agenda to socially coherent behavior. Controlled analyses~\citep{kang2024phyworld} further argue that this world-model framing should remain a benchmark target rather than an assumed capability. Explanation-Oriented Diagnosis then asks whether a system can turn those failures into explicit, interpretable judgments. SPOTLIGHT~\citep{chinchure2025spotlight} and VideoHallu~\citep{li2025videohallu} emphasize error localization and hallucination-aware synthetic-video understanding, while TRAVL~\citep{motamed2025travl} and PhyDetEx~\citep{wang2025phydetex} treat implausibility judgment and explanation as trainable capabilities.
Table~\ref{tab:adjacent-eval} summarizes this adjacent landscape.

\begin{table}[H]
\centering
{\fontsize{6.45}{6.9}\selectfont
\setlength{\tabcolsep}{1.8pt}
\renewcommand{\arraystretch}{0.85}
\begin{tabularx}{\linewidth}{@{}L{0.28\linewidth}L{0.52\linewidth}C{0.10\linewidth}C{0.07\linewidth}@{}}
\toprule
\multicolumn{1}{@{}L{0.28\linewidth}}{\textbf{\textcolor{metablue!90!black}{Work and Resource}}}
& \multicolumn{1}{L{0.52\linewidth}}{\textbf{\textcolor{metablue!90!black}{Description}}}
& \multicolumn{1}{C{0.10\linewidth}}{\textbf{\textcolor{metablue!90!black}{Type}}}
& \multicolumn{1}{C{0.07\linewidth}@{}}{\textbf{\textcolor{metablue!90!black}{Date}}} \\
\midrule
\multicolumn{4}{@{}l}{\textbf{\textcolor{metablue!90!black}{A. Physical Rule Violations}}} \\
\addlinespace[0.10em]
\textbf{VideoPhy}~\citep{bansal2025videophy} & Benchmarks whether generated videos satisfy physical commonsense about objects, attributes, and interactions. & \textit{Eval} & 06/2024 \\
\textbf{Physics-IQ}~\citep{motamed2025generative} & Probes whether text-to-video models obey basic physical principles under controlled prompts. & \textit{Eval} & 01/2025 \\
\textbf{IPV-Bench (Impossible Videos)}~\citep{bai2025impossiblevideos} & Designs impossible scenarios across physical, geographical, biological, and social domains for stress testing generated videos. & \textit{Eval} & 03/2025 \\
\textbf{Morpheus}~\citep{zhang2025morpheus} & Benchmarks physical reasoning of video generative models with real physical experiments and measurable conservation-law violations. & \textit{Eval} & 04/2025 \\
\textbf{T2VPhysBench}~\citep{guo2025t2vphysbench} & Tests first-principles physical consistency in text-to-video generation, including counterfactual robustness and state-transition fidelity. & \textit{Eval} & 05/2025 \\
\textbf{PhyWorldBench}~\citep{gu2025phyworldbench} & Evaluates physical realism of text-to-video models with physics-grounded prompts and anti-physics stress cases. & \textit{Eval} & 07/2025 \\
\textbf{VideoPhy-2}~\citep{bansal2026videophy2} & Extends physical-commonsense evaluation to more challenging action-centric and interaction-heavy generated videos. & \textit{Eval} & 01/2026 \\
\textbf{Physion-Eval}~\citep{zhang2026physioneval} & Provides expert reasoning traces, localized physical glitches, and explanations for physical-realism failures in generated videos. & \textit{Eval} & 03/2026 \\

\addlinespace[0.06em]
\cdashline{1-4}[2pt/2pt]
\addlinespace[0.03em]
\multicolumn{4}{@{}l}{\textbf{\textcolor{metablue!90!black}{B. World Dynamics and Causality}}} \\
\addlinespace[0.10em]
\textbf{WorldSimBench}~\citep{qin2025worldsimbench} & Evaluates whether video generators behave like world simulators by combining perceptual quality and embodied action consistency. & \textit{Eval} & 10/2024 \\
\textbf{Towards World Simulator (PhyGenBench)}~\citep{meng2025towardsworldsimulator} & Crafts a physical-commonsense benchmark for video generation with multi-step dynamics and simulator-style diagnostics. & \textit{Eval} & 10/2024 \\
\textbf{StoryEval}~\citep{wang2024storyeval} & Benchmarks whether T2V models can present short stories composed of consecutive events for future long-video generation. & \textit{Eval} & 12/2024 \\
\textbf{T2VWorldBench}~\citep{chen2025t2vworldbench} & Evaluates world-knowledge generation across physics, nature, activity, culture, causality, and object domains. & \textit{Eval} & 07/2025 \\
\textbf{VideoVerse}~\citep{wang2025videoverse} & World-model-oriented benchmark with hidden semantics, event-level temporal causality, and world-knowledge questions. & \textit{Eval} & 10/2025 \\
\textbf{SVBench}~\citep{peng2025svbench} & Evaluates social reasoning in video generation, including intention, joint attention, norms, and prosocial behavior. & \textit{Eval} & 12/2025 \\

\addlinespace[0.06em]
\cdashline{1-4}[2pt/2pt]
\addlinespace[0.03em]
\multicolumn{4}{@{}l}{\textbf{\textcolor{metablue!90!black}{C. Explanation-Oriented Diagnosis}}} \\
\addlinespace[0.10em]
\textbf{TRAVL}~\citep{motamed2025travl} & Adds intra-frame spatial and trajectory-guided temporal attention and fine-tunes on ImplausiBench to improve VLM judgments of physically implausible videos. & \textit{Method} & 10/2025 \\
\textbf{SPOTLIGHT}~\citep{chinchure2025spotlight} & Fine-grained identification and localization of generation errors using VLMs. & \textit{Eval} & 11/2025 \\
\textbf{VideoHallu}~\citep{li2025videohallu} & Evaluates multimodal hallucination on synthetic videos and studies mitigation via targeted post-training. & \textit{Eval} & 12/2025 \\
\textbf{PhyDetEx}~\citep{wang2025phydetex} & Introduces a benchmark and trains models to detect and explain violated physical rules in text-to-video outputs. & \textit{Method} & 12/2025 \\
\bottomrule
\end{tabularx}}
\caption{Adjacent factual-fidelity diagnostic resources relevant to AIGC-V detector design and world-model-oriented evaluation.}
\label{tab:adjacent-eval}
\end{table}

\FloatBarrier

\section{Challenges and Future}
\label{sec:challenges}

\subsection{Robust Diagnostic Evaluation}
\label{sec:eval-diagnostic}

\citet{chen2024x2dfd} and \citet{sun2025edvd} argue that classification-centric metrics such as clip-level \textit{AUC}/\textit{EER} are often not sufficiently evidential or interpretable for security-sensitive settings.
Under our \emph{factual-fidelity} objective in \S\ref{sec:scope}, strong closed-set discrimination does not necessarily mean the detector can substantiate \emph{which} real-world proposition is violated, \emph{where} the violation occurs, or \emph{why} the judgment is reliable.
Evaluation should therefore be \emph{diagnostic}: it should probe the detector's reliance on complementary pathways (perceptual/temporal cues vs.\ claim-level semantic verification), and reveal the failure modes that lead to factual-fidelity breakdown.

Concretely, one critical axis is robustness of perceptual and temporal evidence under distribution shift.
As diffusion and large video models increasingly suppress noticeable per-frame artifacts, evaluation must stress-test video-level consistency and robustness-to-shift rather than only in-domain performance.
This pressure is already visible in explainable detection settings such as EDVD-LLaMA~\citep{sun2025edvd} and DeeptraceReward~\citep{Fu2025DeeptraceReward}, in generator-diverse detection benchmarks such as GenVidBench~\citep{ni2025genvidbench}, and even in physics-oriented generative evaluation such as Physics-IQ~\citep{motamed2025generative}.
\citet{DBLP:journals/corr/abs-2601-05986} and \citet{hasan2026uneven} further show that strong clean-set scores can mask severe brittleness under transfer-based adversarial attacks and modern synthesis pipelines, leaving a wide gap between benchmark discrimination and deployment robustness.
This motivates cross-dataset generalization protocols, systematic perturbation sweeps over codec, bitrate, and resolution, and deployment-oriented reporting at fixed operating points such as \textit{TPR@FPR=$\alpha$} to better reflect real risk.

A second axis is whether the system can treat a video as making checkable claims and verify them with grounded evidence.
Beyond synchronization, cross-modal alignment tests for retrieval and temporal localization should be complemented with fact-level and knowledge-level verification.
Common sense reasoning for deepfake detection~\citep{zhang2024common} pushes evaluation toward explicit question answering, LVLM-DFD~\citep{Yu2025LVLMDFD} and AuthGuard~\citep{Shen2025AuthGuard} emphasize grounded multimodal reasoning, and BusterX++~\citep{Wen2025BusterXpp} adds unified cross-modal detection-and-explanation settings. These developments also motivate measuring rationale quality and grounding, not just label accuracy.
Together, these diagnostics align evaluation with the shift from perceptual-fidelity discrimination to factual-fidelity verification.

\subsection{Claim-Level and Dynamic Evaluation}
\label{sec:eval-evidence}

A natural next step is to make benchmarks \emph{evidence-first} by moving from clip-level labels to claim-level supervision.
In this setting, each clip is decomposed into a compact set of propositions specifying who, when, where, and what happens, and each proposition is paired with timestamped in-video evidence.
Evaluation then scores both claim correctness and evidence grounding, making ``what is fake'' and ``where it is fake'' traceable rather than implicit.
This direction is consistent with recent explainable AIGC detection resources that augment binary supervision with rationales, traces, and defect-level annotations.
X2-DFD~\citep{chen2024x2dfd} and EDVD-LLaMA~\citep{sun2025edvd} make explanation part of the detector output, Ivy-Fake~\citep{jiang2025ivy} and DAVID-XR1~\citep{gao2025david} attach rationales and defect-level traces to AI-generated video analysis, DeeptraceReward~\citep{Fu2025DeeptraceReward} adds human-perceived fake traces, and ExDDV~\citep{hondru2025exddv} turns explainable supervision into a benchmark design target.

To prevent shortcut learning and isolate genuine reasoning, evidence-first benchmarks should also incorporate targeted stress tests.
One complementary approach is claim-level stress testing for event logic: construct counterfactual or adversarial cases by rewriting event scripts or state transitions while controlling visual-quality confounds, so performance differences primarily reflect relational and event reasoning rather than superficial appearance cues. This concern is explicit in Can We Leave Deepfake Data Behind in Training Deepfake Detector?~\citep{cheng2024can} and FakeChain~\citep{heo2025fakechain}, both of which expose how easily detectors can rely on shallow cues.
In addition, benchmarks increasingly include supervision for tampered locations, spatiotemporal traces, and grounded explanations. EDVD-LLaMA~\citep{sun2025edvd}, DeeptraceReward~\citep{Fu2025DeeptraceReward}, ExDDV~\citep{hondru2025exddv}, and DAVID-XR1~\citep{gao2025david} all move beyond binary labels in this direction, while Video Reality Test~\citep{wang2025videorealitytestaigenerated} adds realistic high-fidelity content where obvious artifacts are weak.
Where available, cross-validating content-side decisions with provenance and authentication signals can further improve trustworthiness in security-sensitive settings.

Finally, given fast-evolving generators, we expect evaluation to move beyond static test sets toward a closed loop between a \emph{real-time detection platform} and a \emph{dynamic benchmark}.
In an arena- or leaderboard-style regime, the benchmark is continuously refreshed with outputs from newly released generators and with realistic ingestion or transcoding effects such as codec and resolution changes; detectors are then periodically re-evaluated under a unified protocol to produce regression trajectories rather than one-off scores~\citep{wang2025videorealitytestaigenerated}. Such a setup measures not only snapshot accuracy, but also robustness decay, adaptation lag, and failure modes under sustained distribution shift. Hard cases that repeatedly fool strong detectors can then be promoted into subsequent benchmark refreshes with failure-type annotations, turning deployment feedback into iterative stress testing rather than passive reporting.

\subsection{Unified Explainable Detection}
\label{sec:dualview-factual}
\citet{wang2025videorealitytestaigenerated} indicates that, for AIGC-V with sparse factual content but high perceptual fidelity and well-aligned modalities, VLMs can perform near random in perceptual-fidelity discrimination, far below humans. This highlights a core gap of language-only detection and the need for visual-view evidence. At the same time, stronger generators and anti-detection pipelines make it increasingly unsafe to assume that authenticity can be decided from perceptual traces alone. Common sense reasoning~\citep{zhang2024common}, generalizable LVLM reasoning~\citep{Yu2025LVLMDFD}, language guidance~\citep{Shen2025AuthGuard}, explainable video reasoning~\citep{gao2025david}, and memory-anchored multimodal reasoning~\citep{chen2025memory} all point toward detectors that must reason over semantic and factual content as well. When perceptual fidelity is high and modalities align, decisive evidence may still come from subtle visual statistics or long-range temporal inconsistencies; frequency-aware blending cues~\citep{zhou2024freqblender}, local dynamic inconsistency learning~\citep{gu2022delving}, natural consistency representations~\citep{zhang2024learning}, and plug-and-play video-level blending with spatiotemporal adapter tuning~\citep{yan2025generalizing} all support this view. Conversely, when generators suppress artifacts and preserve visual coherence, visual-only cues can be insufficient and fact-level reasoning becomes necessary; this is exactly the trajectory emphasized by common-sense and reasoning-heavy detection methods~\citep{zhang2024common,Yu2025LVLMDFD,Shen2025AuthGuard,gao2025david}. Unified detection is therefore a two-pathway problem: perceptual evidence plus fact-level verification.

In practice, this motivates cross-layer evidence fusion: frame-level intrinsic cues from Layer~1, sequence-level motion and physics coherence from Layer~2, multimodal verification from Layer~3, and external-knowledge reasoning from Layer~4 should be integrated into a unified evidence graph, where corroborating evidence boosts confidence and conflicts trigger abstention or human review. The goal, however, is not to average heterogeneous scores, but to preserve evidential granularity so that low-level traces, localized cross-modal conflicts, and claim-level checks remain individually inspectable. The key challenge is to connect low-level cues to high-level conclusions in an auditable way, as X2-DFD~\citep{chen2024x2dfd} and EDVD-LLaMA~\citep{sun2025edvd} already make clear. A practical system should output structured evidence objects, such as suspicious segments and defect-level traces~\citep{gao2025david,anshul2025next}, localized cross-modal mismatches~\citep{katamneni2024contextualcrossmodalattentionaudiovisual,chen2024x2dfd}, and tested claims or entities~\citep{zhang2024common}, rather than only free-form rationales. Tool-augmented pipelines such as LAVID~\citep{Liu2025LAVID} help only when each tool invocation is tied to a concrete sub-claim and yields reusable evidence objects; memory-anchored multimodal reasoning~\citep{chen2025memory} points to the same requirement.

\subsection{Evidence-First Trustworthy Detection}
\label{sec:evidence-first}
To match trustworthy detection that truly targets factual fidelity, system design should follow an Evidence-First principle with an explicit reasoning path: Identify, Localize, and Explain. X2-DFD~\citep{chen2024x2dfd}, DeeptraceReward~\citep{Fu2025DeeptraceReward}, and EDVD-LLaMA~\citep{sun2025edvd} already move in this direction. The key implication is that explanation should be downstream of evidence extraction, not a post-hoc verbal gloss over a classifier score. This path should separate candidate-issue discovery, where-and-when localization, and explanation of which perceptual, temporal, cross-modal, or fact-level constraints are violated. Such a design requires consistent evidence schemas, as DAVID-XR1~\citep{gao2025david} and X2-DFD~\citep{chen2024x2dfd} suggest, together with calibrated uncertainty emphasized by reliability-centered analyses~\citep{Wang2024DeepfakeReliability}. Calibration matters because trustworthy deployment depends on knowing when evidence is weak, conflicting, or incomplete rather than forcing a binary answer.

Within this dual-pathway setup, each tool invocation or knowledge citation should be bound to a specific argumentation step~\citep{Liu2025LAVID}. Content-side analysis should also be cross-validated against source-side provenance and authentication signals when available, rather than treated as an unrelated defense channel. This connection is explicit in media-forensics and defense surveys~\citep{Verdoliva2020MediaForensicsDeepFakes,Deng2025AIGCDefensesSurvey}, in technical provenance standards such as C2PA~\citep{c2pa2024spec}, and in complementary verification frameworks built on face geometry~\citep{tursman2020towards} or blockchain-backed authenticity pipelines~\citep{hajjej2026framework}. This matters beyond standalone detection accuracy: \citet{sagar2026fact} suggests that detector outputs can become misleading priors when they are injected into claim verification pipelines without explicit evidence grounding. The practical difficulty is to reconcile potentially conflicting evidence and surface it in a unified evidence space that supports auditing and abstention under uncertainty~\citep{Wang2024DeepfakeReliability}. A trustworthy system should therefore preserve both agreement and disagreement across evidence sources, rather than collapsing them too early into a single confidence score.

\begin{surveytakeaways}[What Trustworthy Detection Requires]
  \item \textbf{Diagnose the real failure.} Evaluation should reveal where factual fidelity breaks, not only whether real and fake clips separate on a closed set.
  \item \textbf{Supervise evidence, not only labels.} Claim-level annotations, timestamped evidence, shortcut controls, and dynamic testbeds are needed as generators evolve.
  \item \textbf{Return structured evidence, not only scores.} Cross-layer fusion is useful only when cues become traceable claims, segments, entities, or mismatches rather than free-form confidences or rationales.
  \item \textbf{Treat deployment as evidence governance.} Provenance checks, calibration, and abstention are part of trustworthy detection when evidence conflicts or remains incomplete.
\end{surveytakeaways}

\FloatBarrier

\section{Related Work}
\label{sec:relatedwork}

\begin{table}[t]
  \centering
  \footnotesize
  \setlength{\tabcolsep}{1.2pt}
  \renewcommand{\arraystretch}{1.0}
  \begin{tabularx}{\linewidth}{@{}Y*{6}{>{\centering\arraybackslash}m{0.105\linewidth}}@{}}
    \toprule
    & \multicolumn{3}{c}{{\scriptsize\textbf{Scope\,\&\,Framing}}} &
      \multicolumn{3}{c}{{\scriptsize\textbf{Language\,View}}} \\
    \cmidrule(lr){2-4}\cmidrule(lr){5-7}
    \textbf{Survey} &
    \shortstack{\scriptsize Video-\\Targeted\\Scope} &
    \shortstack{\scriptsize Trust-\\worthiness\\Framing} &
    \shortstack{\scriptsize Factual-\\Fidelity\\Verif.} &
    \shortstack{\scriptsize Explain-\\ability\\\& Loc.} &
    \shortstack{\scriptsize Semantic\\Cues} &
    \shortstack{\scriptsize World-\\level\\Reasoning} \\
    \midrule
    \citet{Wang2024DeepfakeReliability} & \xmark & \cmark & \xmark & \xmark & \xmark & \xmark \\
    \citet{Xu2025AIGCForensicsSLR} & \xmark & \cmark & \xmark & \xmark & \xmark & \xmark \\
    \citet{Croitoru2024SurveyOutlook} & \xmark & \cmark & \xmark & \xmark & \xmark & \xmark \\
    \citet{Hashmi2024AVDeepfakeSurvey} & \cmark & \xmark & \xmark & \xmark & \xmark & \xmark \\
    \citet{Kaur2024DeepfakeVideoAIRE} & \cmark & \cmark & \xmark & \xmark & \xmark & \xmark \\
    \citet{Deng2025AIGCDefensesSurvey} & \xmark & \cmark & \xmark & \cmark & \xmark & \xmark \\
    \citet{NguyenLe2025PassiveDF} & \xmark & \cmark & \xmark & \cmark & \xmark & \xmark \\
    \citet{Lin2024DetectingLAIMMultimediaSurvey} & \xmark & \cmark & \xmark & \cmark & \cmark & \xmark \\
    \citet{Zou2025NonMLLMtoMLLMsurvey} & \xmark & \cmark & \xmark & \cmark & \cmark & \xmark \\
    \specialrule{0.3pt}{0pt}{0.3ex}
    \textbf{Ours} & \cmark & \cmark & \cmark & \cmark & \cmark & \cmark \\
    \bottomrule
  \end{tabularx}
  \caption{Comparison of representative surveys and ours across (i) \textbf{Scope and Framing} (video-targeted scope; trustworthiness framing; factual-fidelity verification) and (ii) \textbf{Language View} (explainability and localization; semantic cues; world-level reasoning). \cmark~indicates substantial discussion; \xmark~indicates absent or brief mention.}
  \label{tab:survey_comparison}
\end{table}

Table~\ref{tab:survey_comparison} systematically compares representative surveys by what they take as the task, what deployment concerns they foreground, and how far they move from artifact-centric detection toward grounded verification. This comparison clarifies that our contribution is not only broader coverage, but also a different organizing principle for AIGC-V detection.

\subsection{Scope and Framing}

\paragraph{\textit{Video-targeted scope.}}
Most prior surveys take either a face-manipulation view or a broad \textit{deepfake} or AIGC scope, so they naturally organize the field around generation pipelines, detector families, and generic forensic cues. This breadth is useful for coverage, but it often compresses important distinctions among local manipulation, audio-visual editing, and fully generated video, even though these regimes differ in threat model, evidence availability, and evaluation difficulty. Broad overviews by \citet{Tolosana2020DeepFakesBeyond}, \citet{Mirsky2021CreationDetectionDeepfakesSurvey}, and \citet{Nguyen2022DeepLearningDeepfakesSurvey} are representative of this framing. From a media-forensics viewpoint, \citet{Verdoliva2020MediaForensicsDeepFakes} places deepfakes within a larger integrity-verification pipeline of detection, localization, attribution, and authentication. More video-oriented surveys, including \citet{Hashmi2024AVDeepfakeSurvey} on audio-visual deepfakes and \citet{Kaur2024DeepfakeVideoAIRE} on video-specific challenges, move closer to our focus, but they still do not systematize the full AIGC-V landscape across the three paradigms used here.

\paragraph{\textit{Trustworthiness framing.}}
Recent surveys increasingly move beyond raw accuracy. \citet{Wang2024DeepfakeReliability} and \citet{Deng2025AIGCDefensesSurvey} shift attention toward deployment reliability, adversarial evolution, uncertainty, failure modes, and complementary defenses such as authentication and disruption. Beyond the deepfake-only scope, \citet{Lin2024DetectingLAIMMultimediaSurvey} surveys AI-generated multimedia across modalities and explicitly separates better classification from broader goals such as generalizability, robustness, and interpretability. \citet{Zou2025NonMLLMtoMLLMsurvey} further highlights the transition from domain-specific detectors to VLM- and MLLM-based systems with localization and explanation. Yet even under this framing, trustworthiness is still usually defined around making detector outputs more reliable, rather than around making the underlying authenticity judgment explicitly evidential and claim grounded.

\paragraph{\textit{Factual-fidelity verification.}}
Accordingly, ``verification'' in prior surveys usually appears either inside a broader media-forensics workflow~\citep{Verdoliva2020MediaForensicsDeepFakes} or inside a deployment-reliability discussion centered on robustness, uncertainty, and countermeasures~\citep{Wang2024DeepfakeReliability,Deng2025AIGCDefensesSurvey}. Both perspectives are important, but neither reorganizes the task around what a video claims and whether those claims remain consistent with the world. \citet{Lin2024DetectingLAIMMultimediaSurvey} and \citet{Zou2025NonMLLMtoMLLMsurvey} bring in semantic cues and interpretability with localization, but the emphasis still falls on recognizing manipulations, localizing suspicious regions, or justifying predictions. Fact-level checking of who did what, where, and when is rarely treated as the primary unit of analysis. As summarized in Table~\ref{tab:survey_comparison}, this is the main framing gap that motivates our survey.

\subsection{Language View}

\paragraph{\textit{Explainability \& localization.}}
Recent surveys increasingly recognize explainability and localization, especially in trustworthy and VLM-based detection. \citet{Lin2024DetectingLAIMMultimediaSurvey}, \citet{Zou2025NonMLLMtoMLLMsurvey}, \citet{Deng2025AIGCDefensesSurvey}, and \citet{NguyenLe2025PassiveDF} all acknowledge that deployment requires more than a scalar real/fake score. However, explanation is still usually understood as saliency, manipulated-region localization, or natural-language justification attached to a prediction. What remains less explicit is whether those outputs become reusable evidence objects that can support later claim verification and auditing.

\paragraph{\textit{Semantic cues.}}
As deepfakes move from static artifacts to temporally coherent synthesis, surveys increasingly emphasize video-level evidence and multimodal cues beyond pixels. \citet{Kaur2024DeepfakeVideoAIRE} highlights video-specific challenges and robustness under real-world degradations, while \citet{Hashmi2024AVDeepfakeSurvey} centers audio-visual correspondence, identity preservation, and synchrony cues. \citet{Lin2024DetectingLAIMMultimediaSurvey} and \citet{Zou2025NonMLLMtoMLLMsurvey} push further by acknowledging semantic cues in language-enabled detectors, and \citet{NguyenLe2025PassiveDF} together with \citet{Xu2025AIGCForensicsSLR} consolidate passive detection cues across modalities and emerging AIGC forensics directions. Even so, semantic information is usually introduced as auxiliary context that improves classification or explanation, not as a primary evidential pathway for testing factual fidelity.

\paragraph{\textit{World-level reasoning.}}
Explicit world-level reasoning about events, causality, and plausibility remains even less developed in the survey literature. Most prior surveys stop at semantic alignment or multimodal consistency and do not elevate world knowledge, event logic, or external verification into a first-class evidence pathway. Table~\ref{tab:survey_comparison} makes this asymmetry clear: semantic cues are beginning to appear, but world-level reasoning and, especially, fact-level factual-fidelity verification remain under-emphasized. Our survey addresses this gap by adopting a video-targeted scope, foregrounding factual-fidelity verification, and organizing methods, evaluation, and benchmarks under our taxonomy.

\FloatBarrier

\section{Conclusion}
\label{sec:conclusion}
This survey reframes AI-generated video detection as factual-fidelity verification and synthesizes prior work through a Vision-Language Dual-View four-layer taxonomy with aligned benchmarks and metrics. Beyond summarizing methods, we connect these layers to deployment challenges, evaluation gaps, and emerging trends, and distill key requirements for trustworthy detection, including evidence-first and traceable decisions as well as robustness across generators and real-world scenarios. We expect this survey to provide a clear and actionable reference for future AIGC-V detection research and practice.

More broadly, trustworthy AIGC-V detection should become a shared agenda for the computer vision, natural language processing, multimodal, and world model communities. CV contributes spatiotemporal evidence and forensic robustness; NLP contributes decomposition, reasoning, grounding, and explanation; multimodal and world model research contribute stronger cross-modal alignment and richer priors about physical, causal, and temporal coherence. Together, these ingredients can push detection beyond artifact hunting toward a stricter notion of realism: not whether a video merely looks plausible, but whether its entities, events, and dynamics remain faithful to the constraints of the real world.

\clearpage
\newpage
\bibliographystyle{assets/plainnat}
\bibliography{paper}
\end{document}